\documentclass[pdflatex,sn-mathphys-num]{sn-jnl}% Math and Physical Sciences Numbered Reference Style
\usepackage{graphicx}%
\usepackage{multirow}%
\usepackage{amsmath,amssymb,amsfonts}%
\usepackage{amsthm}%
\usepackage{mathrsfs}%
\usepackage[title]{appendix}%
\usepackage{xcolor}%
\usepackage{textcomp}%
\usepackage{manyfoot}%
\usepackage{booktabs}%
\usepackage{algorithm}%
\usepackage{algorithmicx}%
\usepackage{algpseudocode}%
\usepackage{listings}%
\usepackage{subcaption}
\newcommand{\MyPaperTitle}{Generative emulation of chaotic dynamics with coherent prior}

\title{\MyPaperTitle}

\theoremstyle{thmstyleone}%
\theoremstyle{thmstyletwo}%

\theoremstyle{thmstylethree}%

\begin{document}

%%=============================================================%%
%% GivenName	-> \fnm{Joergen W.}
%% Particle	-> \spfx{van der} -> surname prefix
%% FamilyName	-> \sur{Ploeg}
%% Suffix	-> \sfx{IV}
%% \author*[1,2]{\fnm{Joergen W.} \spfx{van der} \sur{Ploeg} 
%%  \sfx{IV}}\email{iauthor@gmail.com}
%%=============================================================%%

\author*[1,2]{\fnm{Juan} \sur{Nathaniel}}\email{jn2808@columbia.edu}

\author*[1,2]{\fnm{Pierre} \sur{Gentine}}\email{pg2328@columbia.edu}

\affil[1]{\orgdiv{Earth and Environmental Engineering}, \orgname{Columbia University}, \orgaddress{\city{New York}, \postcode{10027}, \state{NY}, \country{USA}}}

\affil[2]{\orgdiv{LEAP NSF Science and Technology Center}, \orgname{Columbia University}, \orgaddress{\city{New York}, \postcode{10027}, \state{NY}, \country{USA}}}

\abstract{Data-driven emulation of nonlinear dynamics is challenging due to long-range skill decay that often produces physically unrealistic outputs. Recent advances in generative modeling aim to address these issues by providing uncertainty quantification and correction. However, the quality of generated simulation remains heavily dependent on the choice of conditioning priors. In this work, we present an efficient generative framework for dynamics emulation, unifying principles of turbulence with diffusion-based modeling: Cohesion. Specifically, our method estimates large-scale coherent structure of the underlying dynamics as guidance during the denoising process, where small-scale fluctuation in the flow is then resolved. These coherent priors are efficiently approximated using reduced-order models, such as deep Koopman operators, that allow for rapid generation of long prior sequences while maintaining stability over extended forecasting horizon. With this gain, we can reframe forecasting as trajectory planning, a common task in reinforcement learning, where conditional denoising is performed once over entire sequences, minimizing the computational cost of autoregressive-based generative methods. Empirical evaluations on chaotic systems of increasing complexity, including Kolmogorov flow, shallow water equations, and subseasonal-to-seasonal climate dynamics, demonstrate Cohesion superior long-range forecasting skill that can efficiently generate physically-consistent simulations, even in the presence of partially-observed guidance.}


%%\pacs[JEL Classification]{D8, H51}

%%\pacs[MSC Classification]{35A01, 65L10, 65L12, 65L20, 65L70}

\maketitle

\section{Introduction}\label{sec1}
Solving partial differential equations (PDEs) with probabilistic emulators \cite{gao2024generative,ruhling2024dyffusion,gilpin2024generative} has gained significant traction over deterministic methods \cite{brandstetter2022clifford,brandstetter2022message,li2020fourier,lu2021learning} because of their ability to generate ensemble forecasts, which are useful for uncertainty quantification and decision-making processes. In particular, diffusion models -- a powerful class of probabilistic models -- have been widely used as emulators in an autoregressive manner to produce sequential outputs over a target lead time, $t \in [0,T]$ \cite{li2024generative,price2023gencast,lippe2024pde}. However, this approach presents several challenges. First, the generative process to estimate the posterior $p(\mathbf{u} \mid \mathbf{c})$ (where $\mathbf{u} \in \mathbb{R}^{T\times n_\mathbf{u}}$ is the state vector and $\mathbf{c} \in \mathbb{R}^{T\times n_\mathbf{c}}$ is the conditioning vector) is highly dependent on the quality of $\mathbf{c}$ \cite{stock2024diffobs,zhao2024advancing,gong2024cascast,gao2024generative,chen2023swinrdm,gao2024prediff,hua2024weather,li2024generative}. Second, diffusion-based autoregressive methods are computationally expensive because they require multiple denoising steps, and the number of function evaluations (NFEs) grows with the discretization granularity of $T$ \cite{price2023gencast,lippe2024pde}. This cost becomes a bottleneck for long-range forecasting applications -- such as in weather and climate -- where previous gains in inference speed achieved by deterministic deep emulators are quickly offset.

\begin{figure}[h]
    \centering
    \includegraphics[width=\textwidth]{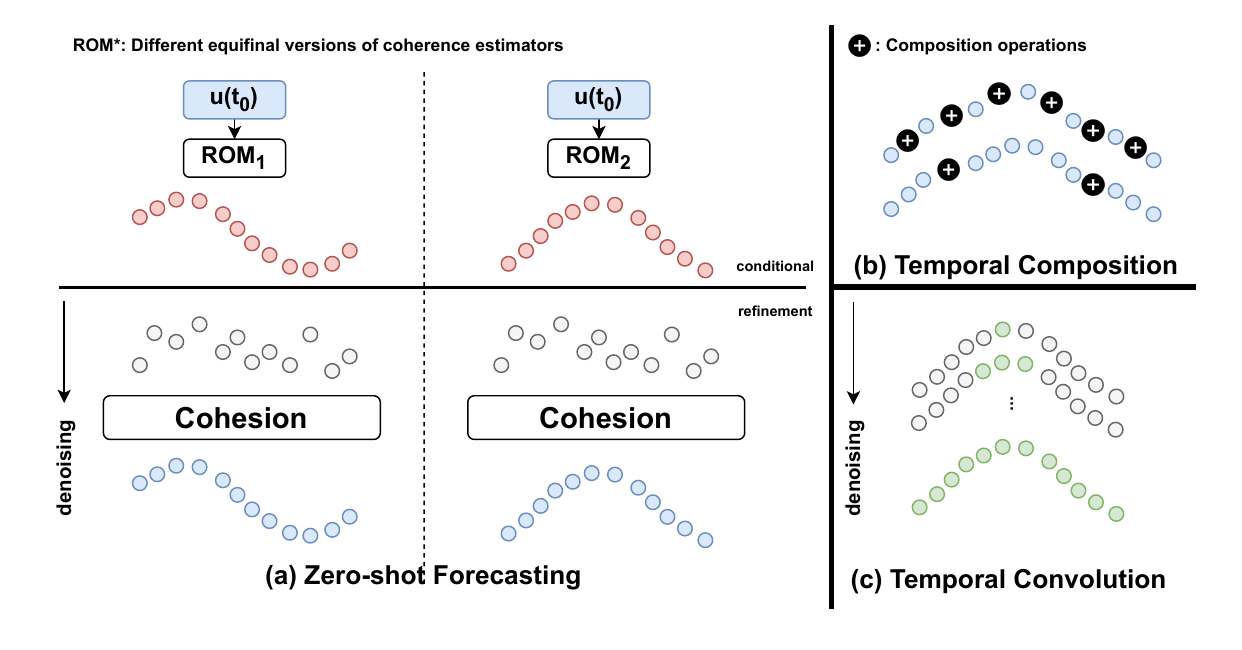}
    \caption{Overview of Cohesion: an emulation framework integrating turbulence principles with diffusion-based modeling. A reduced-order model (ROM) emulates large-scale coherent flow, which serves as a conditioning prior in the generative, resolving process. Inspired by trajectory planning, Cohesion incorporates several key features to enhance flexibility and consistency, including: (a) classifier-free guidance for handling a broad range of conditioning scenarios -- e.g., using different coherent estimators in a ``plug-and-play" manner; (b) temporal composition to accommodate variable-horizon sequences; and (c) model-free convolution to ensure global consistency.}
    \label{fig:cohesion_overview}
\end{figure}

In response to these challenges, we introduce Cohesion (Figure \ref{fig:cohesion_overview}), a diffusion-based emulator that leverages turbulence principles to produce stable and skillful long-range rollouts, while achieving orders-of-magnitude speedups over state-of-the-art baselines. We begin by establishing a theoretical framework in which the generative process in diffusion is interpreted as a closure problem -- that is, one in which the dynamics are expressed as a function of the mean flow -- guided by the coherent prior $\mathbf{c}$. Building upon this framework, we estimate coherent prior using reduced-order models (ROMs), such as the deep Koopman operator \cite{lusch2018deep,wang2022koopman}. ROMs are adept at capturing dynamics that evolve on low-dimensional attractors dominated by persistent coherent structures \cite{stachenfeld2021learned,solera2024beta}, and they offer greater stability over long rollouts compared to high-dimensional models \cite{nathaniel2024chaosbench}. Nonetheless, our framework is agnostic to the choice of coherent estimator, relying on classifier-free guidance that remains invariant to the prior as long as they are stable and effectively capture large-scale dynamics.

By redefining the conditioning vector as a flow, it becomes natural to first generate the coherent trajectory and then perform a conditional pass to resolve the full dynamics in one go. This formulation parallels trajectory planning, a well-studied problem in reinforcement learning (RL) \cite{janner2022planning}, and motivates several techniques to enhance the flexibility and consistency of Cohesion: (i) classifier-free guidance to support a broad range of conditioning scenarios (Figure \ref{fig:cohesion_overview}a), (ii) temporal composition to provide flexibility for variable-horizon emulation (Figure \ref{fig:cohesion_overview}b), and (iii) model-free convolution to improve spatiotemporal consistency (Figure \ref{fig:cohesion_overview}c). 

We evaluate Cohesion by studying chaotic spatiotemporal dynamics of increasing complexity, including Kolmogorov flow (KF), shallow water equation (SWE), and subseasonal-to-seasonal dynamics (S2S; beyond weather timescale, 6-week lead times). For instance, we demonstrate that our framework is both stable and skillful over long rollouts when compared to state-of-the-art probabilistic baselines, including ensemble forecasts from the four leading national weather agencies in the S2S case. Moreover, Cohesion exhibits minimal spectral divergences, highlighting its ability to resolve multiscale physics even in the presence of a partially observed prior. Results are presented in Section \ref{sec2}, with discussion in Section \ref{sec3}, and methodologies detailed in Section \ref{sec4}.

\section{Results }\label{sec2}
% We first introduce our theoretical framework, Cohesion, which incorporates turbulence principles into diffusion-based modeling. We then demonstrate Cohesion's stability and skill in long rollouts, as well as its ability to resolve small-scale physics from coherent priors, even in scenarios with partially observed guidance.  

\subsection{Unified Turbulence-Diffusion Framework}
We recast recent works on diffusion-based dynamics forecasting through the lens of turbulence theory. Specifically, we consider discrete dynamics in one temporal dimension \(t \in [0, T]\):

\begin{equation}
    \label{eq:dynamics}
    \mathbf{u}(t+1) = \mathcal{F}[\mathbf{u}(t)]
\end{equation}

\noindent where \(\mathbf{u}(t) \in \mathbb{R}^{n_\mathbf{u}}\) is the state vector at time $t$, and \(\mathcal{F} : \mathbb{R}^{n_{\mathbf{u}}} \to \mathbb{R}^{n_{\mathbf{u}}}\) differentiable flow map. A common approach to represent these dynamics is to decompose the state $\mathbf{u}$ into a coherent (mean) flow and a fluctuating (turbulent) component. This can be broadly captured by an operator $\mathcal{H} : \mathbb{R}^{T\times n_{\mathbf{u}}} \times \mathbb{R}^{T\times n_{\mathbf{u}}} \to \mathbb{R}^{T \times n_{\mathbf{u}}}$, where the simplest linear case corresponds to the classical Reynolds decomposition:

\begin{equation}
    \label{eq:turbulence}
    \mathbf{u} = \mathcal{H}[\underbrace{\bar{\mathbf{u}}}_{\text{coherent flow}}, \underbrace{\mathbf{u}^\prime}_{\text{fluctuating flow}}] = \underbrace{\bar{\mathbf{u}} + \mathbf{u}^\prime}_{\text{Reynolds decomposition}}
\end{equation}

Often, the turbulent flow is expressed as a function of the mean flow, which is known as the closure problem \cite{vinuesa2022enhancing}. One of the earliest approaches was proposed by Boussinesq \cite{boussinesq1903theorie}, who related Reynolds stresses, $\overline{u^\prime_iu^\prime_j}$ (with the overbar denoting time averages over the $i,j$-th component of $\mathbf{u}^\prime$), to the mean flow through eddy viscosity $\nu_T$. However, this approach -- and many that followed -- assumes a priori model for the unresolved subgrid-scale dynamics \cite{maeyama2020extracting}. 

Recent studies have \textit{implicitly} leveraged this concept in a more data-driven manner, particularly via generative modeling. Here, we make this connection \textit{explicit} starting with the mean flow as our coherent prior, \(\bar{\mathbf{u}}\). Many works, for instance, approximate a variation of \(\bar{\mathbf{u}}\) using a deterministic mapping, which is then used as a conditioning prior to estimate the posterior either (i) directly over the full solution \(p_{\theta}(\mathbf{u}_K \mid \bar{\mathbf{u}})\), or (ii) indirectly over the residual \(p_{\theta}(\mathbf{u}^\prime_K \mid \bar{\mathbf{u}})\) which is subsequently used to correct the mean flow. Throughout this paper, the subscripts $\{0, k, K\}\ \in \mathcal{K}$ refer to the perturbed state vector at the initial, intermediate, and final denoising steps, while the probability density function is generally assumed to be parameterized by $\theta$. We explain each of these strategies in turn and defer detailed preliminaries on diffusion modeling to Section~\ref{sec:diffusion_model}.\\

% \noindent\textbf{Coherent prior for conditioning}. To obtain obtain a conditioning prior within a diffusion framework for forecasting, one typically estimates an initial guess for the current timestep using a parameterized model \(\mathcal{D}: \mathbb{R}^{n_\mathbf{u}} \rightarrow \mathbb{R}^{n_\mathbf{u}}\), such as \(\bar{\mathbf{u}}(\mathbf{x}, t) = \mathcal{D}[\mathbf{u}(\mathbf{x}, t-1)]\) \cite{stock2024diffobs,zhao2024advancing,price2023gencast,gong2024cascast,gao2024generative,chen2023swinrdm}. Others, meanwhile, utilize either a filtered approximation or known system statistics as \(\bar{\mathbf{u}}(\mathbf{x}, t)\) \cite{qu2024deep,gao2024prediff,hua2024weather,li2024generative}. As described earlier, we define deterministic prior to follow closely with the principle of coherent flow in turbulence theory (more in Section \ref{sec:coherence}).\\

\noindent \textbf{Full posterior estimation}. Estimating the full posterior solution involves constructing the operator $\mathcal{H}$ based on the prior approximation and posterior estimation via an iterative denoising process, followed by marginalization over the intermediate states:

\begin{equation}
    \label{eq:diffusion}
    \begin{aligned}
        p_{\theta}(\mathbf{u}_{0:K} \mid \bar{\mathbf{u}}) := p(\mathbf{u}_0)\prod_{k=1}^Kp_{\theta}(\mathbf{u}_k \mid \mathbf{u}_{k-1}, \bar{\mathbf{u}}) \\
        \mathbf{u} \sim p_{\theta}(\mathbf{u}_K \mid \bar{\mathbf{u}}) := \int p_{\theta}(\mathbf{u}_{0:K} \mid \bar{\mathbf{u}}) d\mathbf{u}_{0:K-1}
    \end{aligned} 
\end{equation}

\noindent Subsequently, probability evaluation is performed, for example by taking an expectation over the conditional posterior at the final denoising step $K$ \cite{stock2024diffobs,zhao2024advancing,price2023gencast,gong2024cascast,gao2024generative,chen2023swinrdm,qu2024deep,gao2024prediff,hua2024weather,li2024generative}. \\

\noindent \textbf{Residual posterior estimation}. Several works instead focus to estimate the correction term \(\mathbf{u}'\), rather than the full solution \(\mathbf{u}\) \cite{lippe2024pde,srivastava2023probabilistic,yu2023diffcast,mardani2024residual}. In this approach, $\mathcal{H}$ is first applied as a prior approximation. The posterior estimation step defined in Equation \ref{eq:diffusion} is executed with $\mathbf{u}$ replaced by $\mathbf{u}^\prime$. After marginalizing over the intermediate states and evaluating the posterior, the residual (or stochastic refinement) is added to the prior -- similar to the Reynolds linear decomposition (i.e., the right-hand side of Equation \ref{eq:turbulence}) -- as shown in Equation \ref{eq:diffusion_res} for all $t \in [0, T]$:

\begin{equation}
    \label{eq:diffusion_res}
    \mathbf{u} \approx \underbrace{\bar{\mathbf{u}}}_{\text{deterministic prior}} + \underbrace{\mathbf{u}^\prime}_{\text{stochastic refinement}};\quad \mathbf{u}^\prime \sim p_{\theta}(\mathbf{u}_K^\prime \mid \bar{\mathbf{u}})
\end{equation}

For either approach, variations exist in terms of prior approximation, the sampling/guidance procedure, and additional post-processing steps. Nonetheless, we establish a high-level, conceptual connection between diffusion and turbulence through coherent-prior and fluctuating-refinement pairings. We now proceed to the empirical implementation of this framework, applying it to progressively complex scenarios.

\subsection{Application to Kolmogorov flow}
Incompressible fluid dynamics governed by the Navier-Stokes equations can be defined as:
\begin{equation}
\label{eq:kolmogorov}
\begin{aligned}
    \dot{\mathbf{u}} &= - (\mathbf{u} \cdot \nabla)\mathbf{u} + \frac{1}{\text{Re}} \nabla^2 \mathbf{u} - \frac{1}{\rho} \nabla p + \mathbf{f}, \\
    0 &= \nabla \cdot \mathbf{u}
\end{aligned}
\end{equation}

\noindent where $\mathbf{u}$ is the velocity field, $\text{Re} = 10^3$ is the Reynolds number, $\rho = 1$ is the fluid density, $p$ is the pressure field, and $\mathbf{f}$ is the external forcing. Following \cite{kochkov2021machine} and using $\texttt{jax-cfd}$\footnote{https://github.com/google/jax-cfd} as solvers, we consider a two-dimensional domain \([0, 2\pi]^2\) with periodic boundary conditions and an external forcing \(\mathbf{f}\) corresponding to Kolmogorov forcing with linear damping. We solve the system on a \(256 \times 256\) grid, downsampled to a \(64 \times 64\) resolution, with an integration time step of \(\Delta = 0.2\) model time units between successive snapshots of the velocity field \(\mathbf{u}\). We generate 8196 independent trajectories -- each of length 64 and discarding the first half of warm-ups -- subsequently dividing them into $\texttt{80-10-10}$ train-val-test trajectory-level split. More details on Cohesion setup, probabilistic baselines, and evaluation metrics are described in Sections \ref{sec:experiments} - \ref{sec:metrics}. All models, including Cohesion (score network $s_\theta$ + coherent estimators $f_\psi$), share similar number of trainable parameters for fairness.

\begin{figure}[h]
    \centering
    \includegraphics[width=0.9\textwidth]{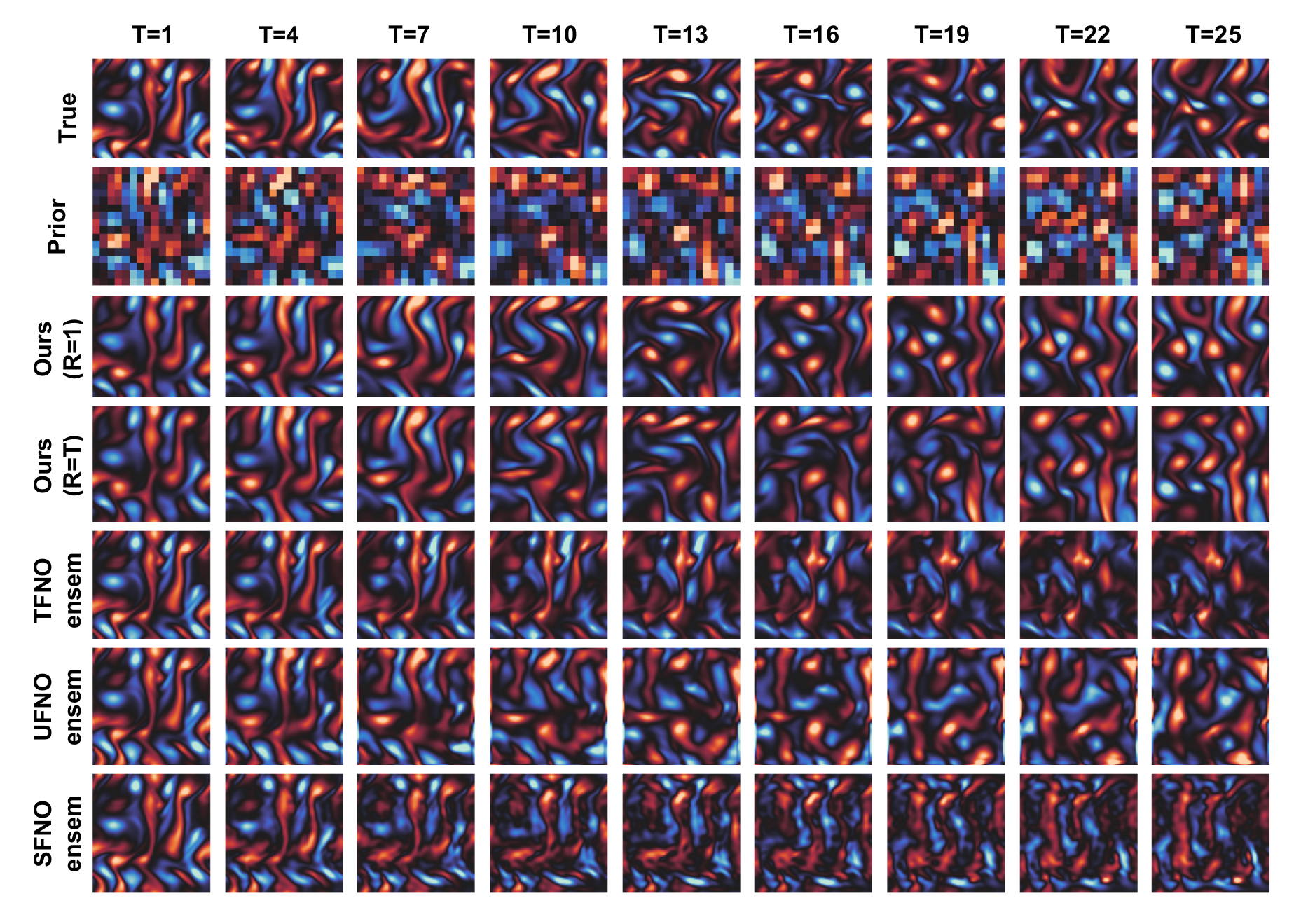}
    \caption{Qualitative single-member realization for Kolmogorov flow where Cohesion is physically-consistent and able to capture fine details over long rollouts compared to its probabilistic baselines.}
    \label{fig:kolmogorov}
\end{figure}

\begin{figure}[h]
    \centering
    \includegraphics[width=\textwidth]{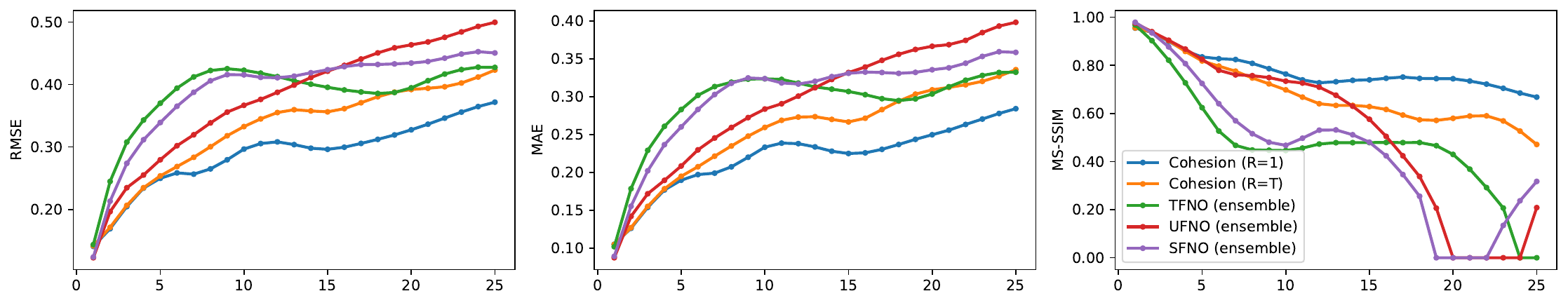}
    \caption{Quantitative ensemble result for Kolmogorov flow where Cohesion has the lowest RMSE ($\downarrow$; Equation \ref{eq:rmse}), MAE ($\downarrow$; Equation \ref{eq:mae}), and highest MS-SSIM ($\uparrow$; Equation \ref{eq:ms-ssim}) over long rollouts compared to its probabilistic baselines.}
    \label{fig:kolmogorov_rollout_metrics}
\end{figure}

As illustrated in Figure \ref{fig:kolmogorov}, we demonstrate that Cohesion produces physically realistic outputs with minimal qualitative instability in both the autoregressive (R=1) and trajectory planning (R=T) settings (see Section \ref{sec:temporal_composition} for details; for brevity trajectory planning performs denoising pass once over the entire forecasting horizon, while autoregressive performs them sequentially). Furthermore, as shown in Figure~\ref{fig:kolmogorov_rollout_metrics}, Cohesion is generally more stable and skillful over long rollouts compared to baseline models, including the probabilistic variants of the Tensorized Fourier Neural Operator (TFNO), UNet-FNO (UFNO), and Spherical FNO (SFNO). To ensure a fair comparison, we conduct extensive ablations to identify the best-performing configurations for each baseline e.g., tuning dropout rates and initial condition perturbation levels, as detailed in Appendix~\ref{secA2}. All results presented in the main text use these optimal hyperparameters. Of particular note is the unstable artifacts for SFNO outputs, which can be attributed to the non-spherical solution of Kolmogorov flow system.

\subsection{Application to shallow water equation}
The SWE system can be described by a set of nonlinear hyperbolic PDEs that governs the dynamics of thin-layer ``shallow" fluid where its depth is negligible relative to the characteristic wavelength. Thus, SWE is ideal to model planetary fluid phenomena \cite{bonev2018discontinuous}. We generate 2048 trajectories of SWE on a rotating sphere as outlined by \cite{bonev2023spherical}, with a $\texttt{80-10-10}$ train-val-test trajectory-level split. Each trajectory is randomly initialized with an average geopotential height of $\varphi_{avg} = 10^3 \cdot g$ and a standard deviation $\varphi_{amp} \sim \mathcal{N}(120, 20) \cdot g$ on a Galewsky setup to mimic barotropically unstable mid-latitude jet \cite{galewsky2004initial}, where $g$ is the gravitational constant. The spatial resolution is $120 \times 240$, keeping the last $T = 32$ of vorticity snapshots. We provide further details on integration scheme, coordinate systems, \textit{etc} in Appendix \ref{si-sec:swe}. Similarly, details on Cohesion setup, probabilistic baselines, and evaluation metrics are described in Sections \ref{sec:experiments} - \ref{sec:metrics}. Similar to Kolmogorov flow, we ensure that all models, including Cohesion (i.e., $s_\theta$ + $f_\psi$) share similar number of trainable parameters for fairness.\\

\begin{figure}[h]
    \centering
    \includegraphics[width=0.9\textwidth]{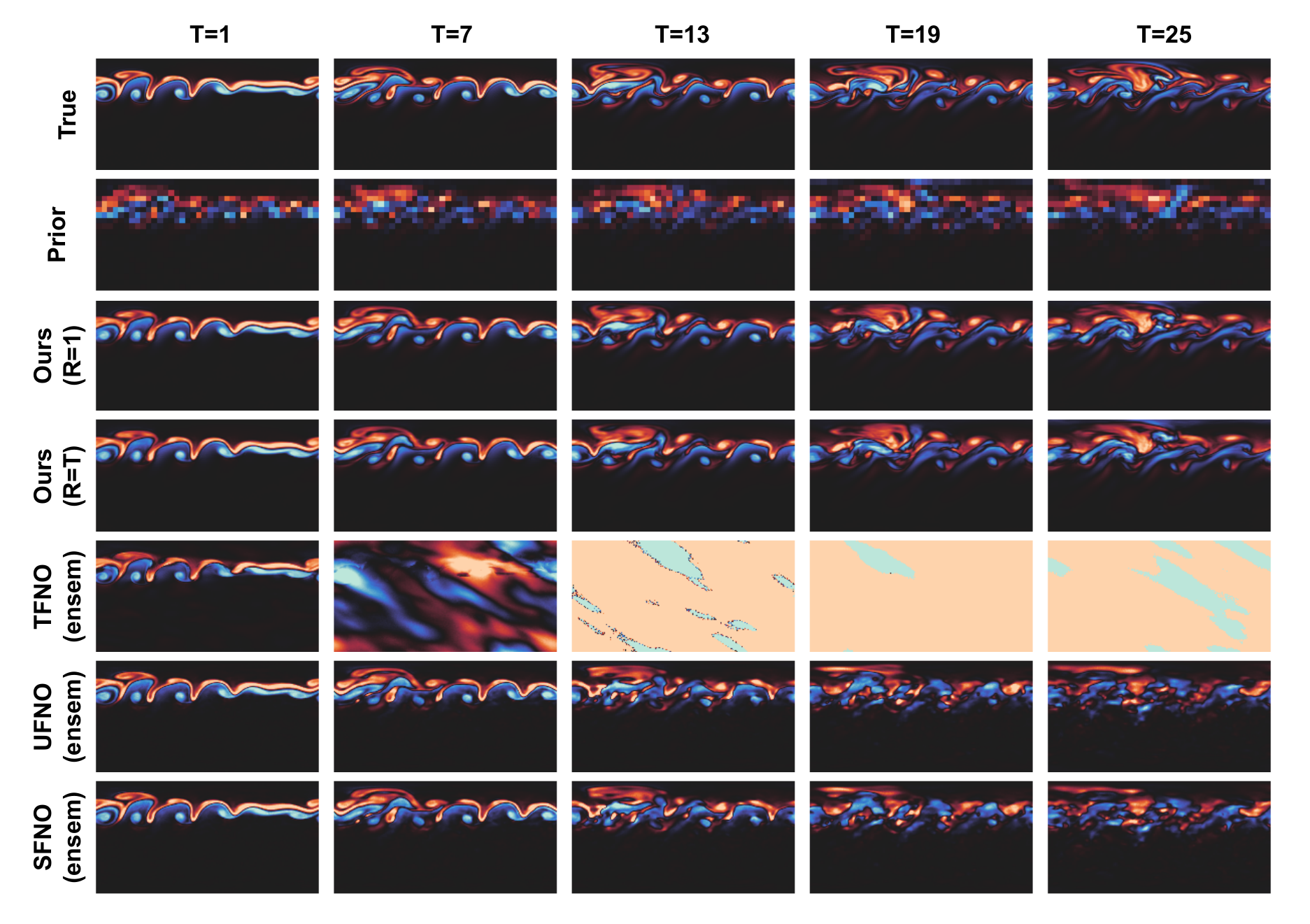}
    \caption{Qualitative single-member realization for Shallow Water Equation where Cohesion is physically-consistent and able to capture fine details over long rollouts compared to its probabilistic baselines.}
    \label{fig:swe}
\end{figure}

\begin{figure}[h]
    \centering
    \includegraphics[width=\textwidth]{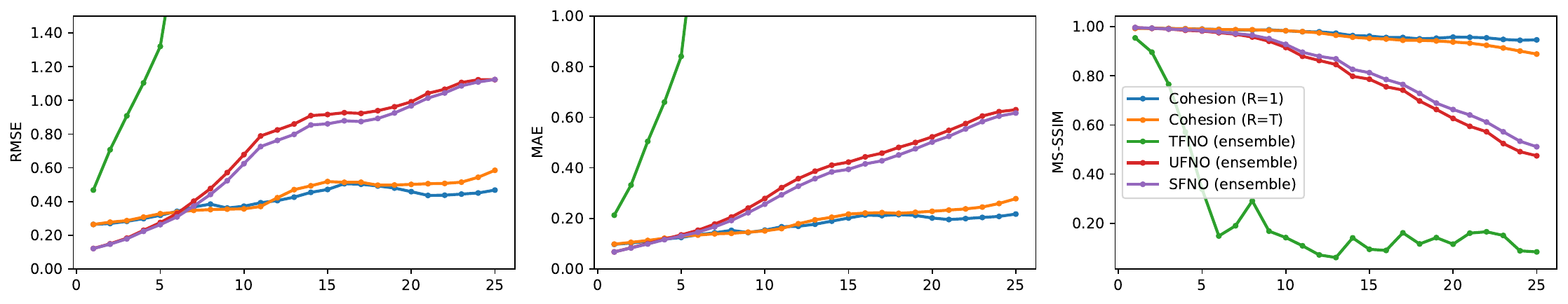}
    \caption{Quantitative ensemble result for Shallow Water Equation where Cohesion has the lowest RMSE ($\downarrow$), MAE ($\downarrow$), and highest MS-SSIM ($\uparrow$) over long rollouts compared to its probabilistic baselines.}
    \label{fig:swe_rollout_metrics}
\end{figure}

As illustrated in Figure~\ref{fig:swe}, Cohesion produces physically realistic outputs with minimal qualitative instability in both the autoregressive (R=1) and trajectory planning (R=T) settings (see Section~\ref{sec:temporal_composition} for details). Furthermore, as shown in Figure~\ref{fig:swe_rollout_metrics}, Cohesion is generally more stable and skillful over long rollouts compared to baseline models, including the probabilistic variants of TFNO, UFNO, and SFNO. Notably, the poor performance of TFNO further underscores the limitations of high-dimensional emulators, particularly in long-range forecasting task where stability is critical \cite{chattopadhyay2023long}.

Before applying Cohesion to emulate real-world S2S climate dynamics, we further evaluate our framework on KF and SWE. We examine its (i) physical consistency, (ii) robustness to incomplete prior, and (iii) scaling behavior including sensitivities to the choice of coherent prior estimators. Unless otherwise stated, all experiments conducted henceforth apply Cohesion in trajectory planning mode (R=T).

\subsection{Physical consistency checks}
 First, we compute the mean spectral divergence (Equation~\ref{eq:specdiv}) over test trajectories. As shown in Figure~\ref{si-fig:specdiv}, Cohesion consistently achieves the lowest divergence, indicating its ability to capture multiscale physical structures (i.e., from low- to high-frequency signals). This is particularly important because strong performance in classical metrics (e.g., RMSE) does not necessarily imply physical consistency \cite{nathaniel2024chaosbench}. As noted earlier, the double penalty issue -- where models tend to predict the mean to avoid placing extremes at incorrect space-time locations -- can inadvertently bias deep learning models toward capturing only large-scale low-frequency patterns. 

\begin{figure}[h]
    \centering
    \begin{subfigure}{0.45\textwidth}
        \centering
        \includegraphics[width=0.9\textwidth]{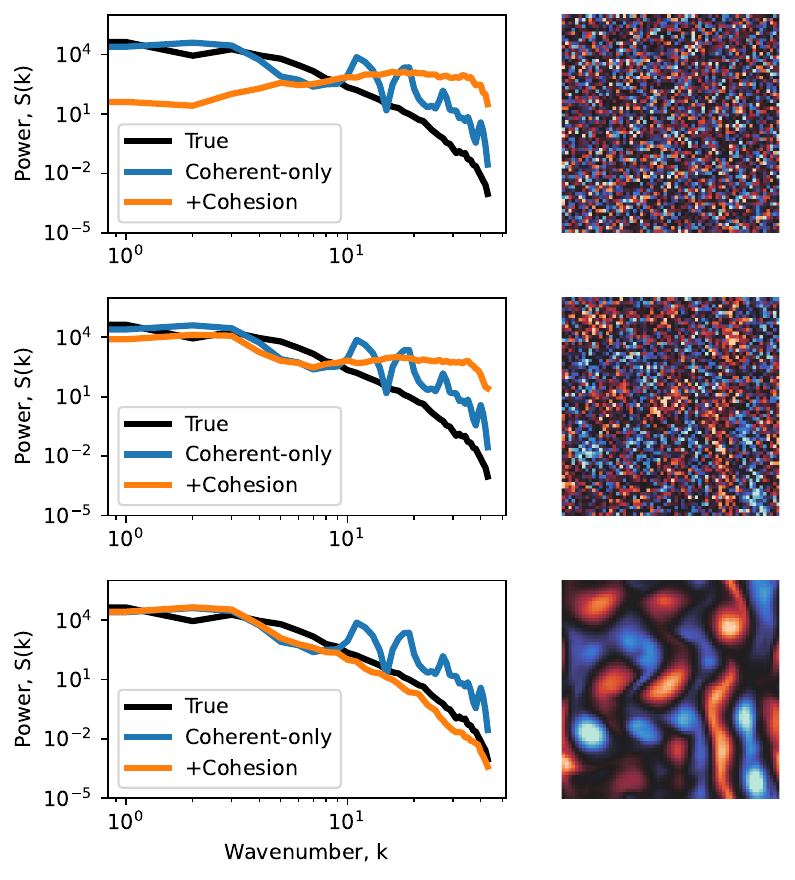}
        \caption{Kolmogorov Flow}
    \end{subfigure}
    \hfill
    \begin{subfigure}{0.45\textwidth}
        \centering
        \includegraphics[width=\textwidth]{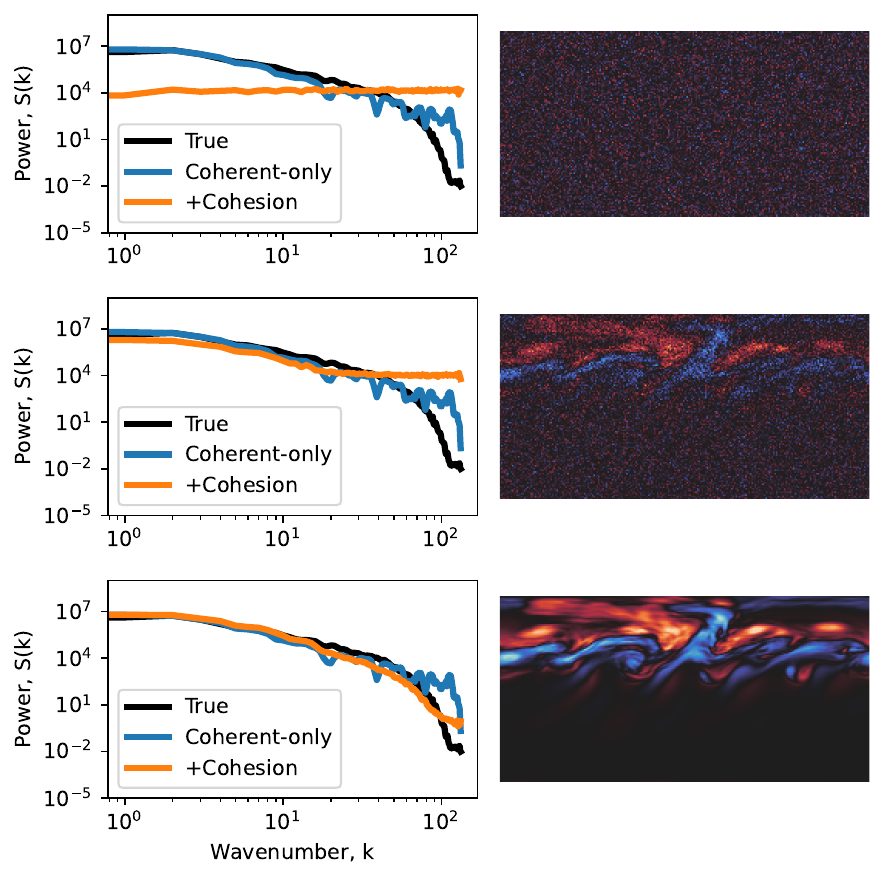}
        \caption{Shallow Water Equation}
    \end{subfigure}
    \hfill 
    
    \caption{Cohesion resolves multiscale physics ubiquitous in chaotic dynamics even after long rollouts ($t=T$), by first getting accurate coherent flow i.e., low-frequency signal ($\downarrow$  wavenumber), before resolving the fluctuating component i.e., high-frequency signal ($\uparrow$ wavenumber). Top-middle-bottom rows represent $\mathbf{u}_k(T)$ at initial-middle-final denoising steps.}
    \label{fig:ablation_closure}
\end{figure}

This result is further illustrated in Figure~\ref{fig:ablation_closure}, where Cohesion successfully resolves multiscale physics by first capturing low-frequency signals (low wavenumber) during the early stages of denoising. These signals correspond to the coherent features of the system (e.g., dominant wave patterns). Once the coherent flow is well-represented, Cohesion proceeds to resolve high-frequency signals (high wavenumber), which are associated with fast-evolving turbulent structures (e.g., eddies) that arise from nonlinear, cascading or coupled interactions. This highlights the capability of Cohesion as a closure mechanism for turbulent flows. Notably, this form of correction also improves other non-spectral metrics, including RMSE, MAE, and MS-SSIM, as shown in Figure~\ref{si-fig:ablation_refiner}.

\subsection{Robustness to incomplete prior}
\begin{figure}[h]
    \centering
    \begin{subfigure}{\textwidth}
        \centering
        \includegraphics[width=0.96\textwidth]{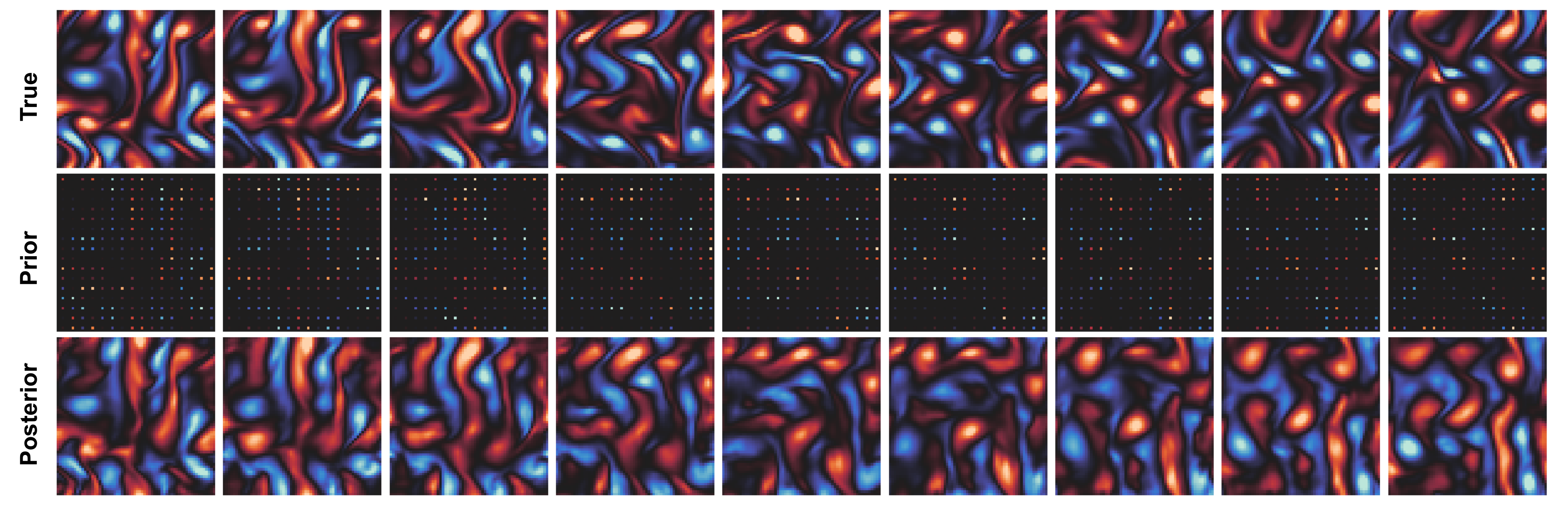}
        \caption{Kolmogorov Flow}
    \end{subfigure}
    \hfill
    \begin{subfigure}{\textwidth}
        \centering
        \includegraphics[width=\textwidth]{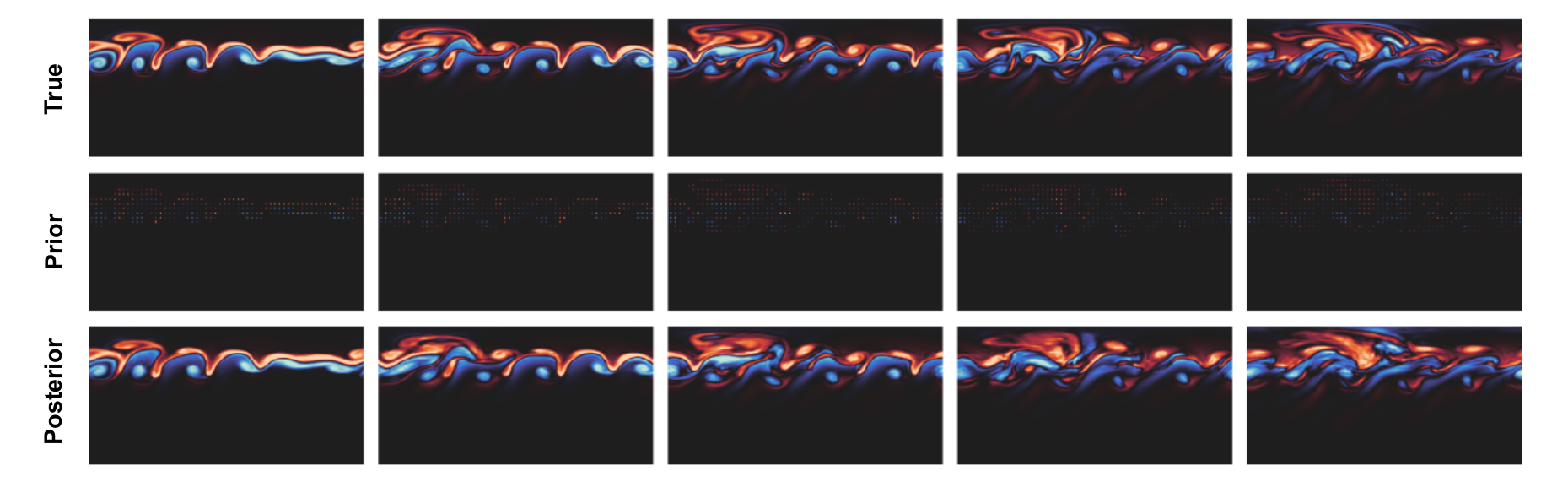}
        \caption{Shallow Water Equation}
    \end{subfigure}
    \hfill 
    
    \caption{Cohesion produces physically-consistent and realistic forecasts at long unrolls even in the presence of partially-observed conditioning prior. In this experiment, we apply equally-spaced masking (every 4 pixels) to the coherent dynamics generated by ROM, which is then used as a conditioning prior during the denoising process.}
    \label{fig:partial_obs}
\end{figure}

Often, the prior of the dynamics we aim to emulate are sparse and noisy (e.g., incomplete signals, partially-observed flow). To evaluate robustness in such settings, we degrade the coherent prior and assess the resulting solution. As shown in Figure~\ref{fig:partial_obs}, Cohesion is able to generate high-resolution, physically realistic outputs even when conditioned on partially observed priors. This property is crucial for modeling real-world systems, where priors may be incomplete (e.g., due to sparse spatial or temporal sampling) \cite{kim2024spatiotemporal}, or inconsistent (e.g., due to systematic model biases) \cite{nathaniel2023metaflux}. It highlights Cohesion's ability to manage uncertainty during long unrolls in a physically grounded manner.

\subsection{Scaling behavior and prior sensitivity tests}
\begin{figure}[h]
    \centering
    \includegraphics[width=0.5\textwidth]{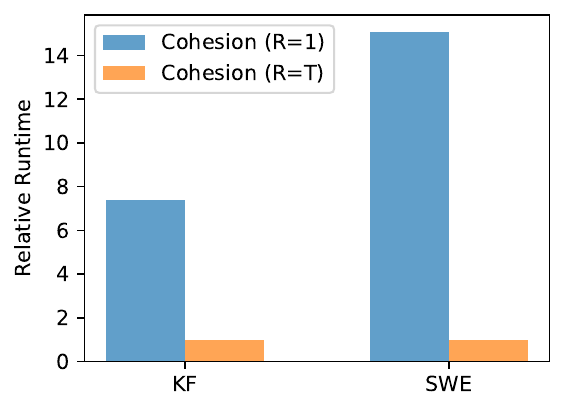}
    \caption{Relative inference runtime in the autoregressive (R=1) and trajectory planning (R=T) settings using identical computational resources. In the latter case, Cohesion achieves orders-of-magnitude speedup compared to traditional autoregressive diffusion-based models, with improvements that scale favorably with the discretization granularity of \(T\).}
    \label{fig:runtime}
\end{figure}

By reframing autoregressive forecasting as trajectory planning, we achieve significant speedup (Figure~\ref{fig:runtime}) without sacrificing performance as demonstrated throughout. These gains are enabled by a combination of reduced-order modeling, temporal composition (see Section \ref{sec:temporal_composition}), and temporal convolution (see Section \ref{sec:temporal_convolution}), which together ensure multiscale consistency across both local and global temporal contexts.

\noindent \textbf{Scaling the dimensionality of coherent prior}. Another informative sanity check is the scaling behavior of Cohesion, particularly with respect to the coherent estimators. We observe that increasing the model size—from small (\(0.25\times\)), to medium (\(0.50\times\)), to large (\(1\times\), default) -- initially improves generative performance, as shown in Figure~\ref{si-fig:ablation_scaling}, but quickly saturates. This suggests diminishing returns beyond a certain model capacity.

\noindent \textbf{Choice of coherent prior estimators}. We also examine the choice of coherent estimators by replacing ROMs with high-dimensional models, including TFNO, UFNO, and SFNO \cite{oommen2024integrating}. In this experiment, the score network \(s_\theta\) is kept fixed, further highlighting the utility of classifier-free guidance in zero-shot forecasting (``plug-and-play"). As shown in Figure~\ref{si-fig:ablation_prior}, poor coherent estimators result in correspondingly poor generative performance. This reinforces our initial hypothesis that the quality of the generated output \(\mathbf{u}\) is highly sensitive to the choice of prior \(\mathbf{c}\), and further motivates the use of coherent estimators that are stable and capable of capturing large-scale dynamics well.

\subsection{Application to subseasonal-to-seasonal climate}
Emulating climate dynamics in the subseasonal-to-seasonal (S2S) scale is particularly challenging due to limited predictability at this timescale, the diminishing leverage from initial condition memory, and the need for robust coupling across Earth system components (e.g., atmosphere–sea ice–land) \cite{vitart2017subseasonal}. Nevertheless, the recent emergence of deep learning weather prediction (DLWP) models, though primarily optimized for medium-range forecasts (5–10 days), has shown promising results and presents fertile ground to extend their predictability range \cite{bi2022pangu,lam2022graphcast,kurth2023fourcastnet}. A key limitation, however, is the blurring effect that arises from the double penalty conundrum \cite{nathaniel2024chaosbench}, where forecasts tend to become overly smoothed over extended rollouts. As a result, such outputs are not physically realistic and are less useful for real-world decision-making tasks, including regional extreme event detection \cite{nathaniel2024inferring}.

Here, we demonstrate the applicability of Cohesion as a probabilistic emulator that (i) enhances the fidelity of smoothed outputs from existing DLWPs through post-processing, and (ii) naturally enables cost-efficient ensembling strategies via generative modeling. For this study, we use the UNet++ architecture\footnote{\href{https://github.com/qubvel-org/segmentation_models.pytorch}{https://github.com/qubvel-org/segmentation\_models.pytorch}} as our coherent prior, extending the work of \cite{weyn2021sub}. It is sufficiently low-dimensional compared to state-of-the-art DLWPs while remaining skillful in capturing large-scale dynamics -- albeit in a smooth manner (see Section~\ref{sec:experiments} for details). Nonetheless, due to our use of classifier-free guidance (``plug-and-play"), Cohesion supports flexible conditioning priors, including outputs from other DLWPs without retraining the score network, though care has to be taken especially in handling long-range drifts \cite{nathaniel2024chaosbench}.

\begin{figure}[h]
    \centering
    \includegraphics[width=\textwidth]{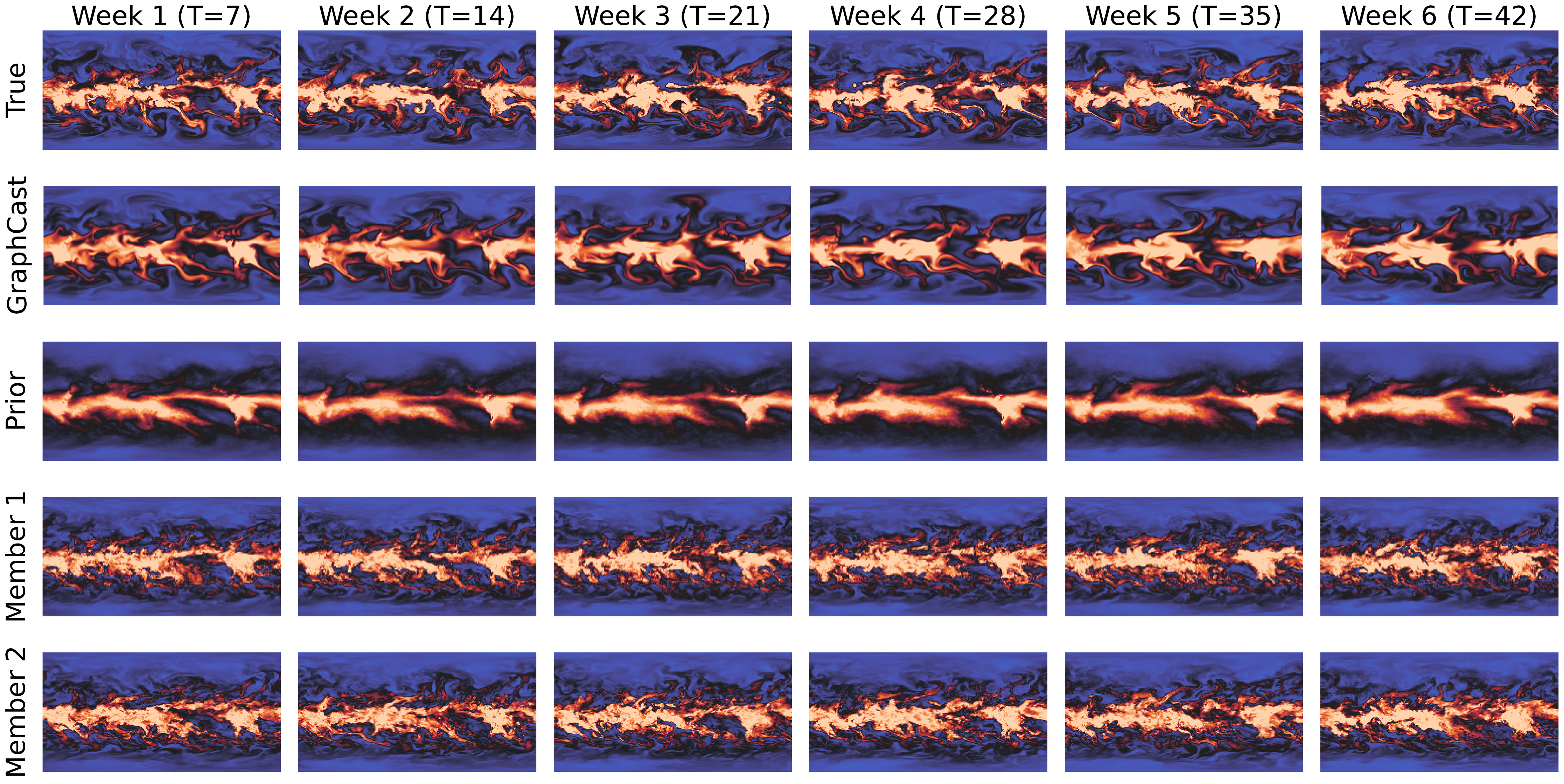}
    \caption{Qualitative realization of S2S dynamics ($q700$), where Cohesion improves the fidelity of forecasts over smoothed-out priors. Initialization date: $2022$-$01$-$01$; results are shown for the simulation snapshot at each of the 6 weeks. We additionally include GraphCast as representative of existing DLWP to highlight the smoothing effect especially at the extratropical region.}
    \label{fig:s2s_q700}
\end{figure}

We benchmark against ensemble forecasts from four national weather agencies: the UK Meteorological Office (UKMO), the National Centers for Environmental Prediction (NCEP), the China Meteorological Administration (CMA), and the European Centre for Medium-Range Weather Forecasts (ECMWF). Additional details are provided in Section~\ref{sec:baselines}. The underlying dataset has a spatial resolution of $1.5^\circ$ ($121 \times 240$ grid) and daily temporal resolution.

\begin{figure}[t!]
    \centering
    \begin{subfigure}{\textwidth}
        \centering
        \includegraphics[width=\textwidth]{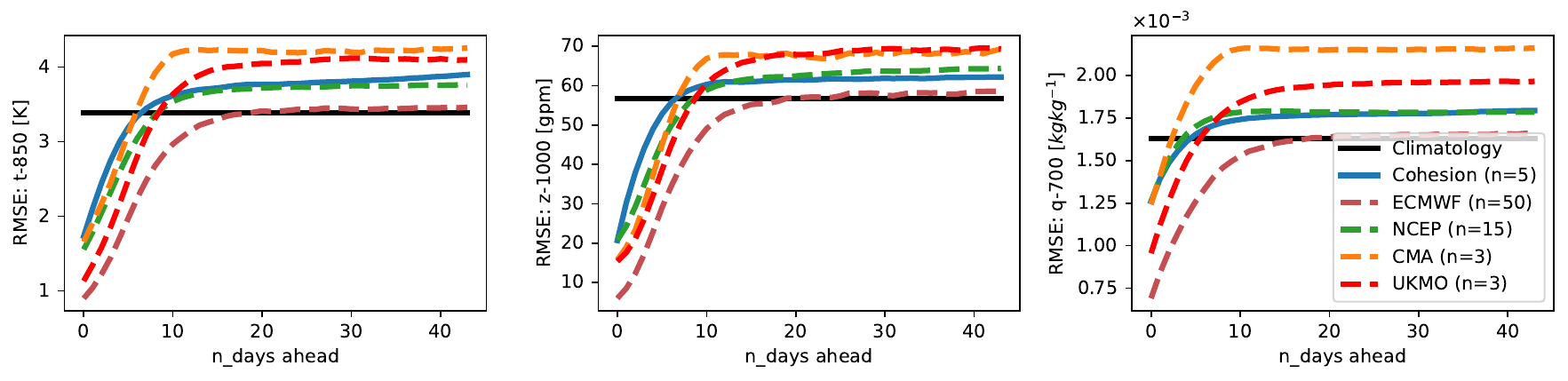}
        \caption{RMSE ($\downarrow$)}
    \end{subfigure}
    \hfill
    \begin{subfigure}{\textwidth}
        \centering
        \includegraphics[width=\textwidth]{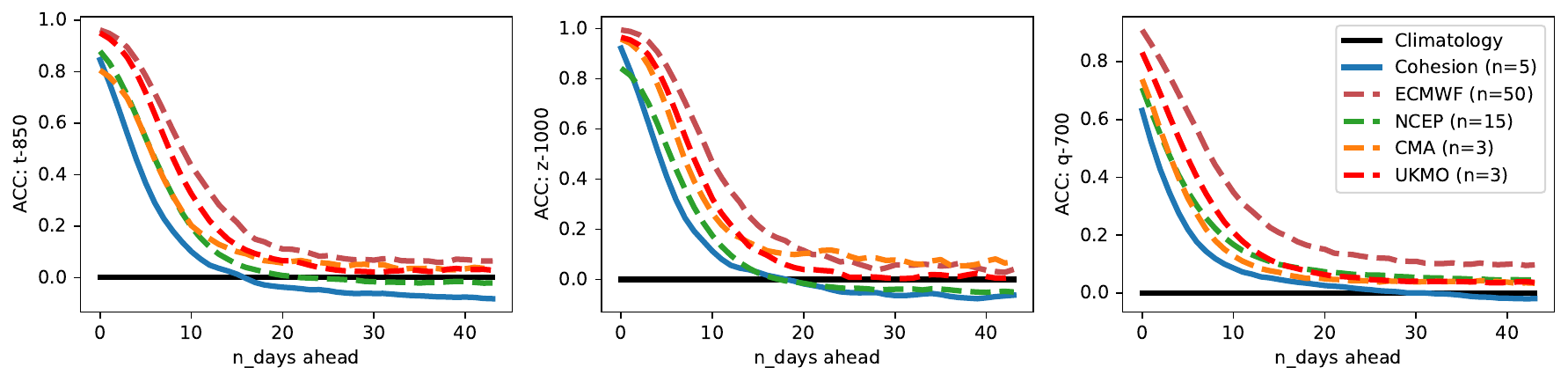}
        \caption{ACC ($\uparrow$)}
    \end{subfigure}
    \hfill 
    \hfill 
    \begin{subfigure}{\textwidth}
        \centering
        \includegraphics[width=\textwidth]{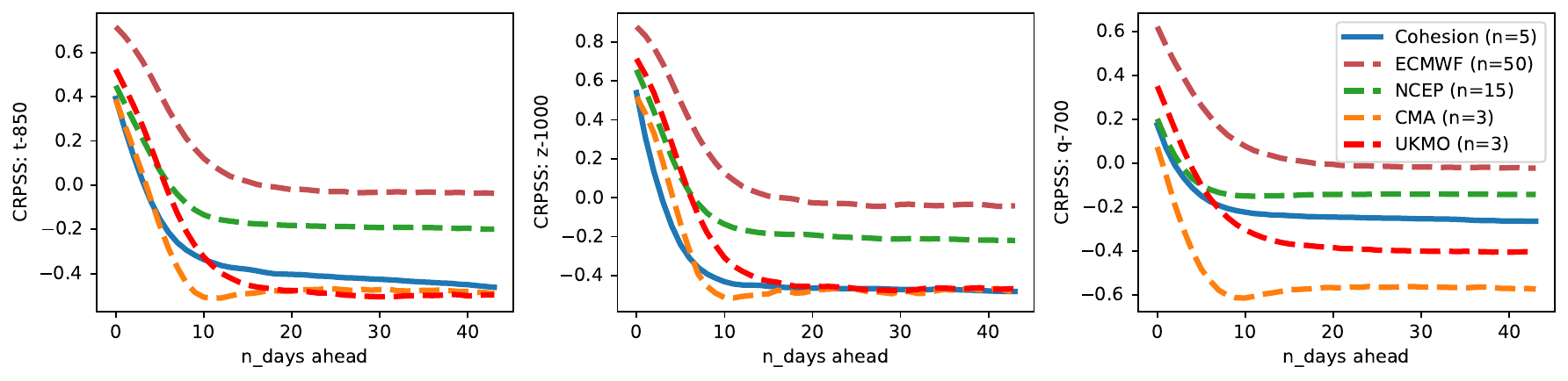}
        \caption{CRPSS ($\uparrow$)}
    \end{subfigure}
    
    \caption{Cohesion is competitive with the ensemble forecasts from ECMWF and NCEP, while often outperforming CMA and UKMO in both deterministic and probabilistic scores over long lead times.}
    \label{fig:s2s_rollout_metrics}
\end{figure}

Training is performed over the years 1979–2019, validation spans 2020–2021, and testing is conducted on 2022–2023. The input variables include geopotential height ($z500$, $z850$, $z1000$), specific humidity ($q700$), temperature ($t850$, $t1000$), and horizontal wind velocity ($u10$, $u200$), with simulations forced by historical sea surface temperature (SST). The numerics after the variable identifiers refer to vertical pressure level in hPa. Evaluations follow the latitude-adjusted metrics from ChaosBench \cite{nathaniel2024chaosbench}, including RMSE, Anomaly Correlation Coefficient (ACC; $\uparrow$), MS-SSIM, Continuous Ranked Probability Score (CRPS; $\downarrow$), and Continuous Ranked Probability Skill Score (CRPSS; $\uparrow$).

As a preliminary illustration, we highlight Figure~\ref{fig:s2s_q700} (also see Figures \ref{si-fig:s2s_t850} - \ref{si-fig:s2s_z1000}) to demonstrate the utility of Cohesion in improving the fidelity of forecasts over smoothed-out priors, serving effectively as a post-processing framework \cite{mouatadid2023adaptive}. To further underscore the persistence of this issue, we compare against GraphCast outputs using the same initialization and forecasting horizon, where the smoothing effect is particularly evident for $q700$ in the extratropical regions.

Furthermore, we evaluate Cohesion's ensemble forecasts and find competitive performance with those of ECMWF and NCEP, often outperforming CMA and UKMO (see Figures~\ref{fig:s2s_rollout_metrics} and~\ref{si-fig:s2s_rollout_metrics}). We also evaluate against a climatology baseline -- an unskilled model that computes the average climate state over the 40-year period beginning in 1979. As expected, skill decay and eventual convergence to climatology are inevitable, as forecasts tend to collapse toward the mean state at long lead times \cite{zhang2019predictability}. Nevertheless, a central goal of Cohesion is to delay this collapse and extend the effective predictability range of any DLWP in a classifier-free manner (``plug-and-play"), while simultaneously improving forecast fidelity and enabling efficient ensembling strategies essential for emulating chaotic systems.

One area of future work may require tighter coupling with other Earth system components, beyond SST forcing used in this work, that retain memory over longer timescales (e.g., soil moisture). Future work should therefore move beyond atmosphere-only emulators -- which remain a standard practice -- and explore the development of coupled systems in order to emulate chaotic dynamics beyond the weather timescale where predictability is especially needed for e.g., disaster management \cite{bauer2021digital}.

\section{Discussion}\label{sec3}
We introduce Cohesion, a diffusion-based emulation framework grounded in turbulence principles that achieves stable, skillful, and physically consistent long rollouts. While emulating chaotic dynamics remains an open challenge, recent advances in deep learning have shown considerable promise. In classical turbulence modeling, for instance, resolving all degrees of freedom is computationally intractable, and thus some level of averaging is typically required \cite{gentine2018could,rasp2018deep}. To approximate the effects of unresolved dynamics, closure models are introduced -- often relying on simplifying assumptions, such as treating small-scale fluctuations as passive and solely dependent on large-scale flow states. However, such assumptions are frequently inadequate, as real-world systems often involve complex multiscale coupling \cite{maeyama2020extracting}.

Generative modeling offers an alternative: instead of assuming the behavior of unresolved dynamics, it learns directly from data. In Cohesion, this is realized through a diffusion-based approach conditioned on coherent flow priors. By explicitly modeling uncertainty and refining predictions through a learned stochastic process, Cohesion serves as a data-driven closure mechanism that resolves fine-scale features without relying on handcrafted parameterizations -- improving fidelity and skill in long-range forecasts. Unlike previous methods, Cohesion explicitly applies turbulence principles which allows for a more theory-grounded, empirically-tested approach to emulate chaotic, multiscale dynamics. In particular, Cohesion establishes a clear connection between coherent flow (as prior), denoising (as closure), and approximation to the full dynamics (as posterior).

Nonetheless, there are several promising research directions. The choice of coherent priors plays a critical role in generative modeling, and designing emulators that can effectively track large-scale dynamics over extended forecasting horizons remains a key challenge. While most ML emulators are optimized for next-step prediction, capturing the correct statistical properties of the system may be equally, if not more important for ensuring long-term stability \cite{schiff2024dyslim}. Our findings further suggest that larger emulators do not necessarily yield better performance; in fact, high dimensionality can often be more of a hindrance than a benefit. This underscores the importance of developing robust ROMs that are both stable and physically grounded, allowing for tighter connection between large-scale coherent dynamics and closure mechanisms provided by generative modeling. Finally, recent advances in generative modeling offer exciting opportunities for turbulence emulation -- from fast diffusion samplers \cite{terpin2024learning} to new generative frameworks such as Schrödinger bridge \cite{de2021diffusion} or flow matching \cite{fotiadis2024stochastic}.

\section{Methods}\label{sec4}
Here, we describe the details of our methods, including the diffusion model specification, baseline emulators, evaluation metrics, and additional experimental details. For notational simplicity and unless otherwise stated, this section refers to $\mathbf{u} \in \mathbb{R}^{n_{\mathbf{u}}}$ as a state vector at a single timestep $t$. 

\subsection{Diffusion Model}
\label{sec:diffusion_model}

We describe our implementation of diffusion model and how zero-shot forecasting is achieved using classifier-free score-based method \cite{song2020score}.\\

\noindent \textbf{Forward diffusion}. At each time step in the forward diffusion process, a sample $\mathbf{u} \sim p(\mathbf{u})$ is progressively perturbed through a continuous diffusion timestepping. This process is described by a linear stochastic differential equation (SDE) as:

\begin{equation}
    \label{eq:forward}
    d\mathbf{u}_k = \underbrace{f(k)\mathbf{u}_k \, dk}_{\text{drift term}} + \underbrace{g(k) \, dw(k)}_{\text{diffusion term}}
\end{equation}

\noindent where $f(k)$ and $g(k) \in \mathbb{R}$ are the drift and diffusion coefficients. Here, $w(k) \in \mathbb{R}^{n_{\mathbf{u}}}$ represents a Wiener process (standard Brownian motion), and $\mathbf{u}_k \in \mathbb{R}^{n_{\mathbf{u}}}$ denotes the perturbed sample at diffusion step $k \in [0,K=1]$. We use \texttt{cosine} noise scheduler in variance-preserving SDE (VPSDE) \cite{nichol2021improved,chen2023importance}.\\

\noindent \textbf{Reverse denoising}. The reverse denoising process is represented by a reverse SDE as defined in Equation \ref{eq:reverse} \cite{song2020score}, where the score function is approximated with a learnable score network, $s_{\theta}(\mathbf{u}_k, k)$. The objective function would be to minimize a continuous weighted combination of Fisher divergences between $s_{\theta}(\mathbf{u}_k, k)$ and $\nabla_{\mathbf{u}_k}\log p(\mathbf{u}_k)$ through score matching \cite{vincent2011connection,song2020score}:

\begin{equation}
    \label{eq:reverse}
    \begin{aligned}
    d\mathbf{u}_k &= [\underbrace{f(k)\mathbf{u}_k}_{\text{drift term}} - g(k)^2 \underbrace{\nabla_{\mathbf{u}_k}\log p(\mathbf{u}_k)}_{\text{score function}}] dk
            + \underbrace{g(k)dw(k)}_{\text{diffusion term}}
    \end{aligned}
\end{equation}

\noindent However, the perturbed state distribution $p(\mathbf{u}_k)$ is data-dependent and unscalable. As such, we reformulate the objective function by replacing $\nabla_{\mathbf{u}_k}\log p(\mathbf{u}_k)$ with $\nabla_{\mathbf{u}_k}\log p(\mathbf{u}_k \mid \mathbf{u})$ where the analytical form of the perturbation kernel is accessible \cite{vincent2011connection}. In order to improve the stability of the objective, especially closer to the start of the denoising step ($k \rightarrow 0$), we apply a reparameterization trick which replaces $s_{\theta}(\mathbf{u}_k, k) = -\epsilon_{\theta}(\mathbf{u}_k, k) / \sigma(k)$, where $\Sigma = \sigma^2$ as in Equation \ref{eq:score_matching_reparam} \cite{zhang2022fast}:

\begin{equation}
    \label{eq:score_matching_reparam}
    \min_{\theta} \mathbb{E}_{p(\mathbf{u}), p(k), p(\epsilon) \sim \mathcal{N}(0, \mathbf{I})}\left[ \left\| \epsilon_{\theta}(\mu(k)\mathbf{u} + \sigma(k)\epsilon, k) - \epsilon) \right\|^2_2 \right]
\end{equation}

% \begin{equation}
%     \label{eq:score_matching}
%     \arg \min_{\theta} \mathbb{E}_{p(\mathbf{u}), p(k), p(\mathbf{u}_k \mid \mathbf{u})}\left[ \Sigma(k)  \left\| s_{\theta}(\mathbf{u}_k, k) - \nabla_{\mathbf{u}_k}\log p(\mathbf{u}_k \mid \mathbf{u}) \right\|^2_2 \right]
% \end{equation}

% Several studies have noted the instability of this objective function nearer to the start of the denoising step ($k \rightarrow 0$), so \cite{zhang2022fast} suggests a reparameterization trick which replaces $s_{\theta}(\mathbf{u}_k, k) = \epsilon_{\theta}(\mathbf{u}_k, k) / \sigma(k)$, where $\Sigma = \sigma^2$ as in Equation \ref{eq:score_matching_reparam}.

\noindent Following standard convention, we denote $\epsilon_{\theta}(\mathbf{u}_k, k)$ with $s_{\theta}(\mathbf{u}_k, k)$ for cleaner notation.\\

\noindent \textbf{Zero-shot conditional sampling}. The case we have discussed so far is the unconditional sampling process as we try to sample $\mathbf{u} \sim p(\mathbf{u}_K)$. In order to condition the generative process with $\mathbf{c} := \bar{\mathbf{u}}$, we seek to sample from $\mathbf{u} \sim p(\mathbf{u}_K \mid \mathbf{c})$. This can be done by modifying the score as in Equation \ref{eq:reverse} with $\nabla_{\mathbf{u}_k}\log p(\mathbf{u}_k \mid \mathbf{c})$ and plugging it back to the reverse SDE process.\\

\noindent As noted earlier, however, one would need fine-tuning whenever the observation process $p(\mathbf{c} \mid \mathbf{u})$ changes. Nonetheless, several works have attempted to approximate the conditional score with just a single pre-trained network, bypassing the need for expensive re-training \cite{song2020score, chung2022diffusion}. Using Bayes rule, we expand the conditional score:

\begin{equation}
    \label{eq:score_bayes}
    \begin{aligned}
    \nabla_{\mathbf{u}_k}\log p(\mathbf{u}_k \mid \mathbf{c}) &= \underbrace{\nabla_{\mathbf{u}_k}\log p(\mathbf{u}_k)}_{\text{unconditional score}} + \underbrace{\nabla_{\mathbf{u}_k}\log p(\mathbf{c} \mid \mathbf{u}_k)}_{\text{log-likelihood function}}
    \end{aligned}
\end{equation}

\noindent Since the first term on the right-hand side is already approximated by the unconditional score network, the remaining task is to identify the second log-likelihood function. Assuming Gaussian observation process, the approximation \cite{chung2022diffusion} becomes:

\begin{equation}
    \label{eq:likelihood_approx}
    \begin{aligned}
    p(\mathbf{c} \mid \mathbf{u}_k) = \int p(\mathbf{c} \mid \mathbf{u}) p(\mathbf{u} \mid \mathbf{u}_k) d\mathbf{u} \approx \mathcal{N}(\mathbf{c} \mid \hat{\mathbf{u}}(\mathbf{u}_k), \sigma^2_\mathbf{c})
    \end{aligned}
\end{equation}

\noindent The mean $\hat{\mathbf{u}}(\mathbf{u}_k)$ can be approximated by the Tweedie's formula \cite{efron2011tweedie}:

\begin{equation}
    \label{eq:tweedies}
    \hat{\mathbf{u}}(\mathbf{u}_k) \approx \frac{\mathbf{u}_k + \sigma^2(k)s_{\theta}(\mathbf{u}_k, k)}{\mu(k)}
\end{equation}

\noindent Following works from \cite{rozet2023score,qu2024deep}, we improve the numerical stability by injecting information about the noise-signal ratio in the variance term, i.e., $\sigma^2_\mathbf{c} + \gamma[\sigma^2(k)/\mu^2(k)]\mathbf{I}$, where $\gamma$, $\mathbf{I}$ are scalar constant and the identity matrix respectively. We now have a classifier-free posterior diffusion sampling where $\nabla{\mathbf{u}_k}\log p(\mathbf{u}_k \mid \mathbf{c})$ can be approximated using a single unconditional score network $s_{\theta}(\mathbf{u}_k, k)$, allowing for zero-shot forecasts given different conditioning scenarios (see Algorithm \ref{alg:posterior_score_estimation}). \\

% \begin{equation}
%     \label{eq:likelihood_reparam}
%     \begin{aligned}
%     p(\mathbf{c} \mid \mathbf{u}_k) \approx \mathcal{N}(\mathbf{c} \mid \hat{\mathbf{u}}(\mathbf{u}_k), \sigma^2_\mathbf{c} + \underbrace{\gamma\frac{\sigma^2(k)}{\mu^2(k)}\mathbf{I}}_{\text{stability term}})
%     \end{aligned}
% \end{equation}

\noindent \textbf{Predictor-corrector}. We implement a predictor-corrector procedure to enhance the quality of our conditional generative process \cite{song2020score}. The reverse SDE prediction process is solved using the exponential integrator (EI) discretization scheme as in Equation \ref{eq:prediction} \cite{zhang2022fast}. The correction phase employs several steps of Langevin Monte Carlo (LMC) to adjust for discretization errors, utilizing a sufficiently small Langevin amplitude $\tau \in \mathbb{R}_+$ as in Equation \ref{eq:correction} (see Algorithm \ref{alg:predictor_corrector}) \cite{song2020score}:

\begin{equation}
    \label{eq:prediction}
    \begin{aligned}
    \mathbf{u}_{k + \Delta k} &\leftarrow \frac{\mu(k + \Delta k)}{\mu(k)}\mathbf{u}_k + \left(\frac{\mu(k + \Delta k)}{\mu(k)} + \frac{\sigma(k + \Delta k)}{\sigma(k)}\right)\Sigma(k)s_{\theta}(\mathbf{u}_k, k \mid \mathbf{c})
    \end{aligned}
\end{equation}

\begin{equation}
    \label{eq:correction}
    \mathbf{u}_k \leftarrow \mathbf{u}_k + \tau s_{\theta}(\mathbf{u}_k, k) + \sqrt{2\tau}\epsilon.
\end{equation}

\subsection{Temporal Composition}
\label{sec:temporal_composition}

When a denoising pass is performed iteratively, as is common in many autoregressive tasks \cite{price2023gencast,srivastava2023probabilistic}, the computational costs becomes prohibitively high. By leveraging compute-efficient estimators for coherent flow, $f_\psi : \mathbb{R}^{n_{\mathbf{u}}} \rightarrow \mathbb{R}^{n_{\mathbf{u}}}$, we can generate a sequence of conditioning priors $\mathcal{C}(R) \in \mathbb{R}^{R \times n_{\mathbf{u}}}$ of length $R$ as follows:

\begin{equation}
    \label{eq:seq_condition}
    \mathcal{C}(R) = \{\mathbf{c}(t_0 + 1) := f^1_\psi(\mathbf{u}(t_0)), \cdots, \mathbf{c}(t_0 + R) := f^R_\psi(\mathbf{u}(t_0))\}
\end{equation}

\noindent where $\mathbf{u}(t_0)$ is the initial condition. We then perform denoising conditioned on $\mathbf{\mathcal{C}}(R)$ to estimate $\mathcal{U}(R) \in \mathbb{R}^{R \times n_{\mathbf{u}}}$ as follows:

\begin{equation}
    \label{eq:seq_solution}
    \mathcal{U}(R) = \{\mathbf{u}(t_0 + 1), \cdots, \mathbf{u}(t_0 + R)\} \sim p_\theta(\mathcal{U}_K(R) \mid \mathcal{C}(R))
\end{equation}

\begin{figure}[h]
    \centering
    \begin{subfigure}{0.49\textwidth}
        \centering
        \includegraphics[width=\textwidth]{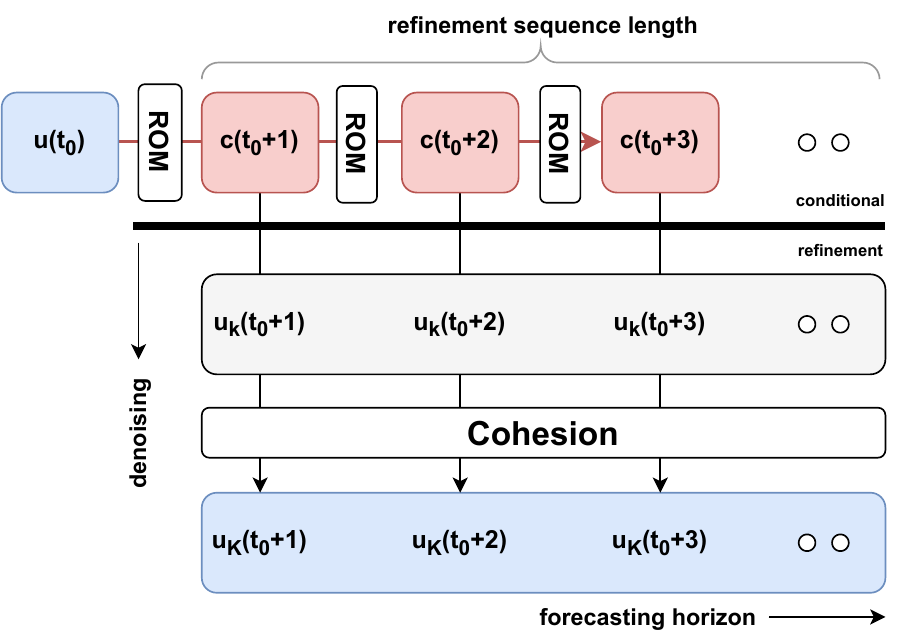}
        \caption{Trajectory planning: $R=T$}
        \label{fig:refine_traj}
    \end{subfigure}
    \hfill
    \begin{subfigure}{0.49\textwidth}
        \centering
        \includegraphics[width=\textwidth]{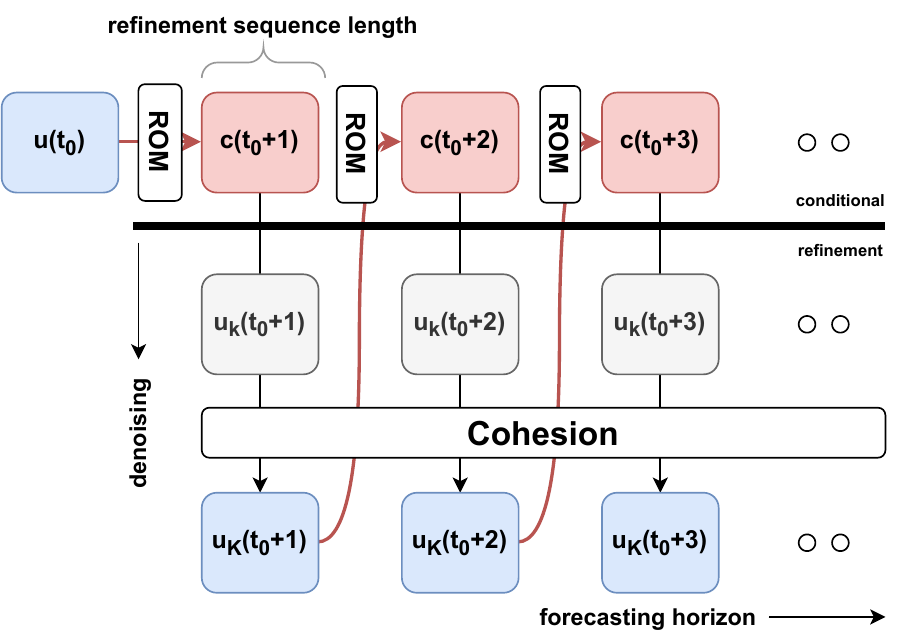}
        \caption{Autoregressive forecasting: $R=1$}
        \label{fig:refine_auto}
    \end{subfigure}
    \hfill 
    
    \label{fig:refine}
    \caption{By temporal composition, we allow for variable-horizon generative emulation with size $R$. (a) Trajectory planning (R=T case) where only one pass of conditional denoising is performed. (b) Autoregressive forecasting (R=1 case) requires multiple denoising passes.}
\end{figure}

\noindent Figure \ref{fig:refine_traj} illustrates an example of trajectory planning ($R = T$) with a single denoising pass. For added flexibility, we allow $R \in [1, T]$. In cases where $R < T$, we can repeat the denoising process, updating the initial condition with the previously generated state vector. Figure \ref{fig:refine_auto} shows this scenario for $R=1$, which corresponds to the classic next-step autoregressive approach.

\subsection{Temporal Convolution}
\label{sec:temporal_convolution}

\begin{figure}[h]
    \centering
    \includegraphics[width=0.7\textwidth]{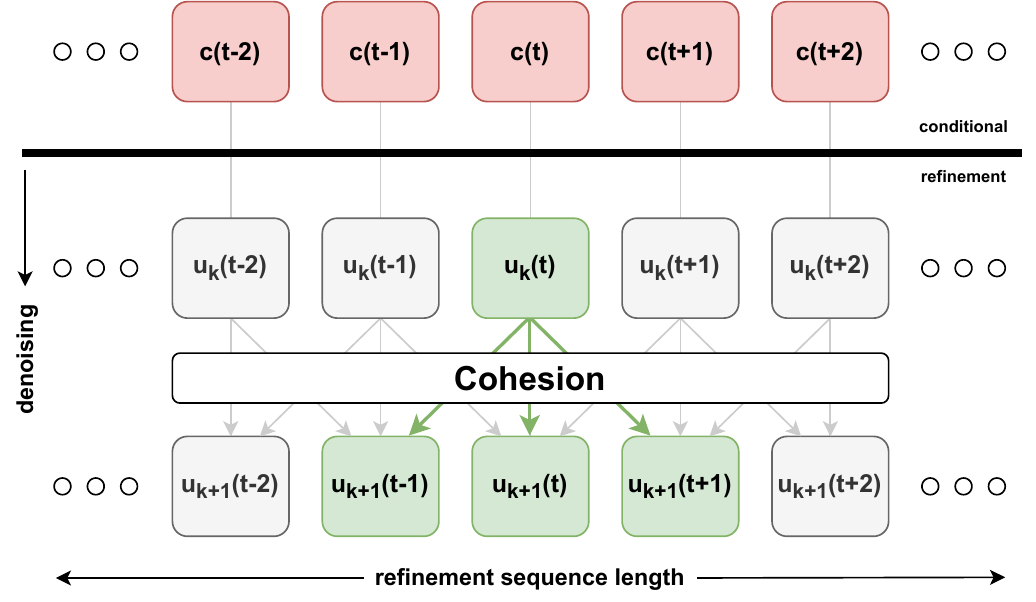}
    \caption{A single denoising step with a local receptive window of size $W = 3$. Global consistency is achieved through multiple compositions across denoising steps, as the score network is both trained and guided over local windows.}
    \label{fig:trajectory}
\end{figure}

To promote temporal consistency, we introduce a model-free local receptive window of size $W \in [1,R]$. This mechanism enables multiscale agreement by guiding the generative process on $W$-length subsequences at each denoising step. The score network is trained to denoise such subsequences, learning local temporal structure. During inference, the same window is applied to maintain consistency across overlapping temporal segments (see Algorithms~\ref{alg:window_training} and~\ref{alg:temporal_convolution} for training and inference procedures using windowed diffusion). Figure~\ref{fig:trajectory} illustrates a single temporal convolution step with a window size of $W=3$ , where local context is incorporated into the refinement. Unless otherwise stated, we use $W=5$ for both training and sampling throughout this work.

\subsection{Experimental Setup}
\label{sec:experiments}
We describe additional experimental setups, including the choice of coherent estimators and final hyperparameter configurations. \\

\noindent \textbf{Kolmogorov flow and shallow water equation}. For our coherent estimators, we use deep Koopman operators. Koopman theory \cite{koopman1932dynamical} posits that nonlinear dynamics can be modeled by an infinite-dimensional linear Koopman operator acting on the space of all possible observables. 
Leveraging a deep encoder-decoder model to learn these observables, $\{\mathcal{G}_E, \mathcal{G}_D\} \in \mathcal{G}$ \cite{lusch2018deep}, the Koopman operator $\mathcal{O}: \mathbb{R}^{n_\mathbf{d}} \rightarrow \mathbb{R}^{n_\mathbf{d}}$
acts on a lower ($n_\textbf{d}$)-dimensional latent manifold that advances the state vector in time:

\begin{equation}
    \label{eq:koopman_encode}
    \mathcal{O}[\mathcal{G}_E(\mathbf{u}(t))] := \mathcal{G}_E \circ \mathcal{F}[\mathbf{u}(t)] = \mathcal{G}_E \circ \mathbf{u}(t + 1)
\end{equation}

\noindent A conditioning prior is then generated by the decoder as: 

\begin{equation}
    \label{eq:koopman_decode}
    \bar{\mathbf{u}}(t + 1) := \mathcal{G}_D \circ \mathcal{O}[\mathcal{G}_E(\mathbf{u}(t))]
\end{equation}

\noindent Composing $\mathcal{O}$ for $m$ times within Equation \ref{eq:koopman_decode} results in the generation of an autoregressive sequence of conditioning priors that extends over $m$ steps. We perform joint training by minimizing the 1-step lagged reconstruction loss as in Equation \ref{eq:koopman_loss}. We collectively refer to $\{\mathcal{G}_E, \mathcal{O}, \mathcal{G}_D\} \in f_\psi$ as the coherent estimators, where \(f_\psi : \mathbb{R}^{n_{\mathbf{u}}} \to \mathbb{R}^{n_{\mathbf{u}}}\). 

\begin{equation}
    \label{eq:koopman_loss}
    \min_{\{\mathcal{G}_E, \mathcal{O}, \mathcal{G}_D\} \in f_\psi} \mathbb{E}_{p(\mathbf{u})}\left[ \left\| \bar{\mathbf{u}}(\mathbf{x}, t+1) - \mathbf{u}(\mathbf{x}, t+1) \right\|^2_2 \right]
\end{equation}

\noindent For KF, $f_\psi$ consists of 6 symmetrical convolution layers in the $\mathcal{G}_E - \mathcal{O} - \mathcal{G}_D$ composition with hidden size of [4, 8, 16, 32, 64, 128] and embedding dimension of $n_\textbf{d} = 128$. The score network, $s_\theta$ is parameterized by modern U-Net with [3, 3, 3] residual blocks \cite{he2016deep}, each consisting of [16, 32, 64] hidden channels. The temporal component of $s_\theta$ is parameterized by a two-layer dense network with 256 hidden channels and 64-dimensional embedding. Both models, $f\psi$ and $s_\theta$, are trained  over 256 epochs, optimized with \textsc{AdamW} \cite{loshchilov2017decoupled}, with a batch size of 64, learning rate of $2\times10^{-4}$, and a weight decay of $1\times10^{-3}$. During inference, we apply 64 denoising steps with 1-step LMC correction, $\gamma=10^{-2}$, and $\tau=3e^{-2}$. Training and inference are performed using 1x A100 NVIDIA GPU in a 100GB memory node.\\

\noindent For SWE, $f_\psi$ consists of 5 symmetrical convolution layers in the $\mathcal{G}_E - \mathcal{O} - \mathcal{G}_D$ composition with hidden size of [8, 16, 32, 64, 128] and embedding dimension of $n_\textbf{d} = 128$. The score network, $s_\theta$ is parameterized by modern U-Net with [3, 3, 3] residual blocks \cite{he2016deep}, each consisting of [16, 32, 64] hidden channels. The temporal component of $s_\theta$ is parameterized by a two-layer dense network with 256 hidden channels and 64-dimensional embedding. Both models, $f\psi$ and $s_\theta$, are trained  over 256 epochs, optimized with \textsc{AdamW} \cite{loshchilov2017decoupled}, with a batch size of 64, learning rate of $2\times10^{-4}$, and a weight decay of $1\times10^{-3}$. During diffusion inference, we apply 64 denoising steps with 1-step LMC correction, $\gamma=10^{-2}$, and $\tau=3e^{-2}$. Training and inference are performed using 1x A100 NVIDIA GPU in a 100GB memory node.\\

\noindent \textbf{Subseasonal-to-seasonal climate dynamics}. For S2S, we estimate $f_\psi$ using UNet++ \cite{zhou2018unet++}. We use \texttt{resnet34} as our backbone, with a symmetrical convolution blocks of sizes [32, 64, 128] and embedding dimension of $n_{\mathbf{d}} = 128$. The score network, $s_\theta$ is parameterized by modern U-Net with [3, 3, 3] residual blocks \cite{he2016deep}, each consisting of [32 64, 128] hidden channels. The temporal component of $s_\theta$ is parameterized by a two-layer dense network with 256 hidden channels and 64-dimensional embedding. Both models, $f\psi$ and $s_\theta$, are trained  over 128 epochs, optimized with \textsc{AdamW} \cite{loshchilov2017decoupled}, with a batch size of 16, learning rate of $1\times10^{-4}$ and $1\times10^{-2}$, and a weight decay of $1\times10^{-2}$. During diffusion inference, we apply 32 denoising steps with 1-step LMC correction, $\gamma=10^{-2}$, and $\tau=0.5$. Training and inference are performed using 2x A100 NVIDIA GPU in a 150GB memory node.\\

\noindent \textit{Bias correction}. We also implement bias correction for each field using the strategy proposed by \cite{weyn2021sub}. However, rather than applying correction to both the control and ensemble reforecasts, we correct only the control in order to highlight Cohesion's capacity for further refinement if needed. Specifically, we generate forecasts initialized daily using the coherent prior (control run) on a fixed set of calendar dates spanning 1979–2019, covering the training period and part of the validation set. For each date, a spatially varying bias is computed as the average bias over the 40-year period. During inference on the test set, we subtract the corresponding bias from the coherent forecast for that date, and use this debiased field for guidance (see Figure \ref{si-fig:s2s_bias_correct}).\\

\noindent \textit{Timescale separation}. Furthermore, we train two coherent prior estimators to capture both \texttt{fast} and \texttt{slow} time scales. The fast model is trained by minimizing over 2-day autoregressive rollouts, while the slow model is trained similarly but over 10 days. During inference, we linearly weigh outputs from the two models using a decaying weight over the 0–10 day horizon, and rely solely on the slow model beyond day 10.

\subsection{Probabilistic Baselines}
\label{sec:baselines}

\noindent \textbf{Kolmogorov flow and shallow water equation}. We use Spherical Fourier Neural Operator (SFNO) \cite{bonev2023spherical}, tensorized FNO (TFNO) \cite{li2020fourier}, and U-Shaped FNO (UFNO) \cite{rahman2022u} as our baseline, implemented off-the-shelves from \href{https://github.com/NVIDIA/torch-harmonics}{https://github.com/NVIDIA/torch-harmonics} and \href{https://neuraloperator.github.io}{https://neuraloperator.github.io}. To ensure a fair comparison, we scale all baselines parameters to match or exceed those of Cohesion and introduce probabilistic modifications. Since these models are not probabilistic by default, we perform probabilistic transformation as described in \cite{ruhling2024dyffusion}. Unless stated, all models are evaluated on five samples/members.

\begin{itemize}
    \item \textit{Checkpoints}: Ensembles through multiple model fitting initialized randomly
    \item \textit{MC-Dropout}: Ensembles by enabling inference-time dropouts
    \item \textit{IC Perturbation}: Ensembles through the perturbation of initial conditions
\end{itemize}

\noindent \textbf{Subseasonal-to-seasonal climate dynamics}. We use ChaosBench \cite{nathaniel2024chaosbench} to benchmark Cohesion. In particular, we evaluate against S2S ensemble outputs from four leading national agencies across the world, with data freely accessible from \href{https://huggingface.co/datasets/LEAP/ChaosBench}{https://huggingface.co/datasets/LEAP/ChaosBench}. 

\begin{itemize}
    \item \textit{UKMO Ens}: The UK Meteorological Office \cite{williams2015met}
    \item \textit{NCEP Ens}: National Centers for Environmental Prediction \cite{saha2010ncep}
    \item \textit{CMA Ens}: China Meteorological Administration \cite{wu2019beijing}
    \item \textit{ECMWF Ens}: European Center for Medium-Range Weather Forecasts \cite{documentationpart}
\end{itemize}

\subsection{Evaluation Metrics}
\label{sec:metrics}

We group our metrics into pixel-, structure-, and physics-based. The first two quantify information loss in the physical space, while the latter in the spectral space.\\

\noindent \textbf{Pixel-based metrics}. We use root mean-squared error (RMSE; Equation \ref{eq:rmse}) and mean absolute error (MAE; Equation \ref{eq:mae}), as follows:

\begin{equation}
    \label{eq:rmse}
    \mathcal{M}_{RMSE} = \sqrt{\frac{1}{n_{\textbf{u}}} \sum (\hat{\mathbf{u}} - \mathbf{u})^2}
\end{equation}

\begin{equation}
    \label{eq:mae}
    \mathcal{M}_{MAE} = \frac{1}{n_\mathbf{u}} \sum |\hat{\mathbf{u}} - \mathbf{u}|
\end{equation}

\noindent \textbf{Structure-based metrics}. Let $\mathbf{Y}$ and $\hat{\mathbf{Y}}$ be two images to be compared, and let $\mu_\mathbf{Y}$, $\sigma^2_\mathbf{Y}$ and $\sigma_{\mathbf{Y}\hat{\mathbf{Y}}}$ be the mean of $\mathbf{Y}$, the variance of $\mathbf{Y}$, and the covariance of $\mathbf{Y}$ and $\hat{\mathbf{Y}}$, respectively. The luminance, contrast and structure comparison measures are defined as follows:
\begin{equation}
    l(\mathbf{Y}, \hat{\mathbf{Y}})=\frac{2 \mu_\mathbf{Y} \mu_{\hat{\mathbf{Y}}}+C_1}{\mu_\mathbf{Y}^2+\mu_{\hat{\mathbf{Y}}}^2+C_1},
\end{equation}
\begin{equation}
    c(\mathbf{Y}, \hat{\mathbf{Y}})=\frac{2 \sigma_\mathbf{Y} \sigma_{\hat{\mathbf{Y}}}+C_2}{\sigma_\mathbf{Y}^2+\sigma_{\hat{\mathbf{Y}}}^2+C_2},
\end{equation}
\begin{equation}
    s(\mathbf{Y}, \hat{\mathbf{Y}})=\frac{\sigma_{ \mathbf{Y}\hat{\mathbf{Y}}}+C_3}{\sigma_\mathbf{Y} \sigma_{\hat{\mathbf{Y}}}+C_3},
\end{equation}
where $C_1$, $C_2$ and $C_3$ are constants given by:
\begin{equation}
C_1 = (K_1L)^2, C_2 = (K_2L)^2, \text{ and } C_3 = C_2/2.
\end{equation}
$L=255$ is the dynamic range of the gray scale images, and $K_1\ll 1$ and $K_2 \ll 1$ are two small constants. To compute the MS-SSIM metric across multiple scales, the images are successively low-pass filtered and down-sampled by a factor of 2. We index the original image as scale 1, and the desired highest scale as scale $M$. At each scale, the contrast comparison and structure comparison are computed and denoted as $c_j(\mathbf{Y},\hat{\mathbf{Y}})$ and $s_j(\mathbf{Y},\hat{\mathbf{Y}})$ respectively. The luminance comparison is only calculated at the last scale $M$, denoted by $l_M(\mathbf{Y},\hat{\mathbf{Y}})$. Then, the MS-SSIM metric is defined by:
\begin{equation}
\label{eq:ms-ssim}
\mathcal{M}_{MS-SSIM} =  [l_M(\mathbf{Y}, \hat{\mathbf{Y}})]^{\alpha_M}\cdot\prod_{j=1}^M[c_j(\mathbf{Y}, \hat{\mathbf{Y}})]^{\beta_j}[s_j(\mathbf{Y}, \hat{\mathbf{Y}})]^{\gamma_j}
\end{equation}
where $\alpha_M$, $\beta_j$ and $\gamma_j$ are parameters. We use the same set of parameters as in \cite{wang2003multiscale}: $K_1 = 0.01$, $K_2 = 0.03$, $M=5$, $\alpha_5=\beta_5=\gamma_5=0.1333$, $\beta_4=\gamma_4=0.2363$, $\beta_3=\gamma_3=0.3001$, $\beta_2=\gamma_2=0.2856$, $\beta_1=\gamma=0.0448$. The predicted and ground truth images of physical variables are re-scaled to 0-255 prior to the calculation of their MS-SSIM values.\\

\noindent \textbf{Physics-based metrics}. We next describe the definition and implementation of our physics-based metrics, particularly Spectral Divergence (SpecDiv). Consider a 2D image field of size $h \times w$ for a physical parameter at a specific time, variable, and level. Let $f(x,y)$ be the intensity of the pixel at position $(x,y).$  First, we compute the 2D Fourier transform of the image by:

\begin{equation}
    F(k_x, k_y) = \sum_{x=0}^{w-1}\sum_{y=0}^{h-1}f(x,y)\cdot e^{-2\pi i \left(k_xx/w+ k_yy/h\right)}
    \label{eq:DFT}
\end{equation}

\noindent where $k_x$ and $k_y$ correspond to the wavenumber components in the horizontal and vertical directions, respectively, and $i$ is the imaginary unit. The power at each wavenumber component $(k_x, k_y)$ is given by the square of the magnitude spectrum of $F(k_x, k_y)$, that is:

\begin{equation}
    S(k_x, k_y) = \vert F(k_x, k_y) \vert ^2 = \texttt{Re}[F(k_x, k_y)]^2 + \texttt{Im}[F(k_x, k_y)]^2
\end{equation}

\noindent The scalar wavenumber is defined as:

\begin{equation}
    k=\sqrt{k_x^2 + k_y^2}
    \label{eq:k}
\end{equation}

\noindent which represents the magnitude of the spatial frequency vector, indicating how rapidly features change spatially regardless of direction. Then, the energy distribution at a spatial frequency corresponding to k is defined as:

\begin{equation}
    S(k)=\sum_{(k_x,k_y):\sqrt{k_x^2+k_y^2}=k}S(k_x, k_y)
    \label{eq:Ek}
\end{equation}

\noindent Given the spatial energy frequency distribution for observations $S(k)$ and predictions $\hat{S}(k)$ , we perform normalization over $\mathbf{K}$, the set of wavenumbers, as defined in Equation \ref{eq:norm_ek}. This is to ensure that the sum of the component sums up to 1 which exhibits pdf-like property. 

\begin{equation}
    S(k) \gets \frac{S(k)}{\sum_{k\in\mathbf{K}}S(k)}, \quad
    \hat{S}(k) \gets \frac{\hat{S}(k)}{\sum_{k\in\mathbf{K}}\hat{S}(k)}
    \label{eq:norm_ek}
\end{equation}

\noindent Finally, the SpecDiv is formalized as follows:

\begin{equation}
    \label{eq:specdiv}
    \mathcal{M}_{SpecDiv} = \sum_{k} S(k) \cdot \log(S(k) / \hat{S}(k))
\end{equation}

\noindent where $S(k), \hat{S}(k)$ are the power spectra of the target and forecast in space continuum.\\

\noindent \textbf{Other metrics}. As discussed in the main text, additional metrics for evaluating S2S forecasts -- including latitude-weighted variants and probabilistic scores such as CRPS and CRPSS -- are detailed in ChaosBench \cite{nathaniel2024chaosbench}.

\backmatter

\section*{Data availability}
Synthetic data is generated according to the script provided in \href{https://github.com/juannat7/cohesion}{https://github.com/juannat7/cohesion}. S2S data is public and freely accessible at \href{https://huggingface.co/datasets/LEAP/ChaosBench}{https://huggingface.co/datasets/LEAP/ChaosBench}.

\section*{Code availability}
Code will be open sourced at this link upon paper publication: \href{https://github.com/juannat7/cohesion}{https://github.com/juannat7/cohesion}.

\section*{Acknowledgements}
The authors acknowledge funding,
computing, and storage resources from the NSF Science and Technology Center (STC) Learning
the Earth with Artificial Intelligence and Physics (LEAP) (Award \#2019625). JN also acknowledge funding from Columbia-Dream Sports AI Innovation Center.

\section*{Ethics declarations}
The authors declare no competing interests.

\clearpage

\bibliography{main}

\clearpage

\begin{appendices}

\setcounter{figure}{0}
\renewcommand{\thefigure}{S\arabic{figure}}

\setcounter{table}{0}
\renewcommand{\thetable}{S\arabic{table}}

\setcounter{equation}{0}
\renewcommand{\theequation}{S\arabic{equation}}

\setcounter{algorithm}{0}
\renewcommand{\thealgorithm}{S\arabic{algorithm}}

\begin{center}
    {\Large \bfseries Supplementary Materials}\\
    {\large \MyPaperTitle}\\
    \textit{Juan Nathaniel and Pierre Gentine}
\end{center}

\section{Problem Setup}\label{secA0}
We provide additional details on experimental setups.

\subsection{Shallow Water Equation}
\label{si-sec:swe}
We define a new coordinate system on the spherical domain $\mathbf{x} \in \mathcal{S}^2$ in terms of longitude $\varphi \in [0, 2\pi]$ and colatitude $\theta \in [0,\pi]$. The unit vector $\mathbf{x}$ can then be reparamterized as $(\cos(\varphi)\sin(\theta), \sin(\varphi)\sin(\theta), \cos(\theta))^\intercal$. Given this coordinate transform, we define the set of differential equations describing SWE:

\begin{equation}
    \label{eq:swe}
    \begin{aligned}
        \partial_t\varphi + \nabla\cdot(\varphi u) = 0\\
        \partial_t(\varphi u) + \nabla\cdot T = \mathcal{S}
    \end{aligned}
\end{equation}

\noindent with initial conditions $\varphi = \varphi_0, u = u_0$. The state vector $(\varphi, \varphi u^\intercal)^\intercal$ includes both the geopotential layer depth $\varphi$ (representing mass) and the tangential momentum vector $\varphi u$ (indicative of discharge). Within curvilinear coordinates, the flux tensor $T$ can be expressed using the outer product $\varphi u \otimes u$. The right-hand side of the equation features flux-related terms, such as the Coriolis force. We use spectral method to solve the PDE on an equiangular grid with a spatial resolution of $120 \times 240$ and 60-second timesteps. Time-stepping is performed using the third-order Adams-Bashford scheme and snapshots are taken every 5 hour for a total of 12 days, keeping the last 32 temporal sequences of vorticity outputs. The parameters of the PDE, such as gravity, radius
of the sphere and angular velocity, are referenced to the Earth.

\clearpage

\section{Pseudocode and Algorithms}\label{secA1}
We present algorithms for some of the processes described in main text.

\begin{algorithm}[H]
\caption{Posterior score estimation, $\nabla_{\mathbf{u}_k}\log p(\mathbf{u}_k \mid \mathbf{c})$}
\label{alg:posterior_score_estimation}
    \begin{algorithmic}[1]

        \Function{PosteriorEstimate}{$s_\theta, \mathbf{u}_k, k, \mathbf{c}$}
        \State $s_\mathbf{u} \gets s_\theta(\mathbf{u}_k, k)$
        \State $\hat{\mathbf{u}} \gets \frac{\mathbf{u}_k + \sigma^2(k)s_\mathbf{u}}{\mu(k)}$
        \State $s_\mathbf{c} \gets \nabla_{\mathbf{u}_k}\log\mathcal{N}(\mathbf{c} \mid \hat{\mathbf{u}}, \sigma^2_{\mathbf{c}} + \gamma\frac{\sigma^2(k)}{\mu^2(k)}\mathbf{I})$
        \State \Return $s_\mathbf{u} + s_\mathbf{c}$
        \EndFunction
    \end{algorithmic}
\end{algorithm}

\begin{algorithm}[H]
\caption{Predictor-corrector sampling}
\label{alg:predictor_corrector}
    \begin{algorithmic}[1]

        \Function{DiffusionSampling}{$s_\theta, \tau, N_c$}
        \State $\mathbf{u}_{\mathcal{K}_{0} = 0} \sim \mathcal{N}(0, \sigma^2(\mathbf{I}))$
        \For{$i=0$ to $|\mathcal{K}|$}
            \State $s_p \gets \mathtt{PosteriorEstimate}(s_\theta, \mathbf{u}_{\mathcal{K}_{i}}, \mathcal{K}_{i}, \mathbf{c})$ \Comment{see Algorithm \ref{alg:posterior_score_estimation}}
            \State $\mathbf{u}_{\mathcal{K}_{i+1}} \leftarrow \frac{\mu(\mathcal{K}_{i+1})}{\mu(\mathcal{K}_{i})}\mathbf{u}_{\mathcal{K}_{i}} + \left(\frac{\mu(\mathcal{K}_{i+1})}{\mu(\mathcal{K}_{i})} - \frac{\sigma(\mathcal{K}_{i+1})}{\sigma(\mathcal{K}_{i})}\right)\sigma^2(\mathcal{K}_{i})s_p$ \Comment{Predictor}
            
            \For{$j=0$ to $N_c$}
                \State $\epsilon \sim \mathcal{N}(0, \mathbf{I})$
                \State $s_c \gets \mathtt{PosteriorEstimate}(s_\theta, \mathbf{u}_{\mathcal{K}_{i+1}}, \mathcal{K}_{i+1}, \mathbf{c})$  \Comment{see Algorithm \ref{alg:posterior_score_estimation}}
                \State $\mathbf{u}_{\mathcal{K}_{i+1}} \leftarrow \mathbf{u}_{\mathcal{K}_{i+1}} + \tau s_c + \sqrt{2\tau}\epsilon$  \Comment{Corrector}
            \EndFor
        \EndFor
        \State \Return $\mathbf{u}_K$ 
        \EndFunction
    \end{algorithmic}
\end{algorithm}

\begin{algorithm}[H]
\caption{Score network training with window of size, $W$}
\label{alg:window_training}
    \begin{algorithmic}[1]
        \Require $W \mod 2 = 1$ \Comment{Symmetric window about $\mathbf{u}(t_i)$}
        \State $W \gets 2w + 1$
        \While{not done}
            \State $i \sim \mathcal{U}(\{w+1, \cdots, T-w\})$
            \State $k \sim \mathcal{U}(\mathcal{K}), \epsilon \sim \mathcal{N}(0, \mathbf{I})$
            \State $\mathbf{u}_k(t_{i -w : i + w}) \gets \mu(k)\mathbf{u}(t_{i -w : i + w}) + \sigma(k)\epsilon$
            \State $\texttt{Loss} \gets \left\| \epsilon_{\theta}(\mathbf{u}_k(t_{i -w : i + w}), k) - \epsilon) \right\|^2_2$
            \State $\theta \gets \texttt{GradientUpdate}(\theta, \nabla_{\theta}\texttt{Loss})$
        \EndWhile
        
    \end{algorithmic}
\end{algorithm}

\begin{algorithm}[H]
\caption{Temporal convolution with local receptive window within (sub)sequences}
\label{alg:temporal_convolution}
    \begin{algorithmic}[1]
        \Require $1 \leq w \leq R$
        \Function{TemporalConvolution}{$s_\theta, \mathbf{u}_k, \mathbf{c}, k, w, R$}
        \State $s_{1:w+1} \gets s_\theta(\mathbf{u}_k(t_{1:2w+1}), k \mid \mathbf{c})[:w+1]$
        \For{$i = w + 2$ to $R - w - 1$}
            \State $s_i \gets s_\theta(\mathbf{u}_k(t_{i-w:i+w}), k \mid \mathbf{c})[w+1]$
        \EndFor
        \State $s_{R-w:R} \gets s_\theta(\mathbf{u}_k(t_{R-2w:R}), k \mid \mathbf{c})[w + 1:]$
        \State \Return $s_{1:R}$
        \EndFunction
        
    \end{algorithmic}
\end{algorithm}

\clearpage

\section{Ablation Study}\label{secA2}
We perform extensive ablation in order to choose the best baseline configurations. For instance, which ensembling strategies and the corresponding hyperparameters yield the best emulators.

\subsection{Ensembling: MC-dropout}
We vary the probability of dropout, $p$, during inference (MC-dropout) in order to construct an ensemble forecasts. \\

\begin{figure}[h]
    \centering
    \begin{subfigure}{\textwidth}
        \centering
        \includegraphics[width=\textwidth]{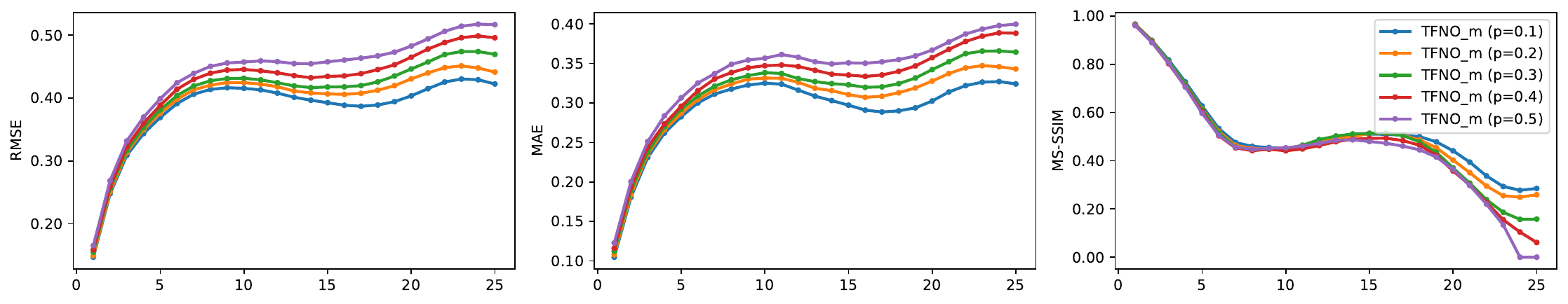}
        \caption{TFNO}
    \end{subfigure}
    \hfill
    \begin{subfigure}{\textwidth}
        \centering
        \includegraphics[width=\textwidth]{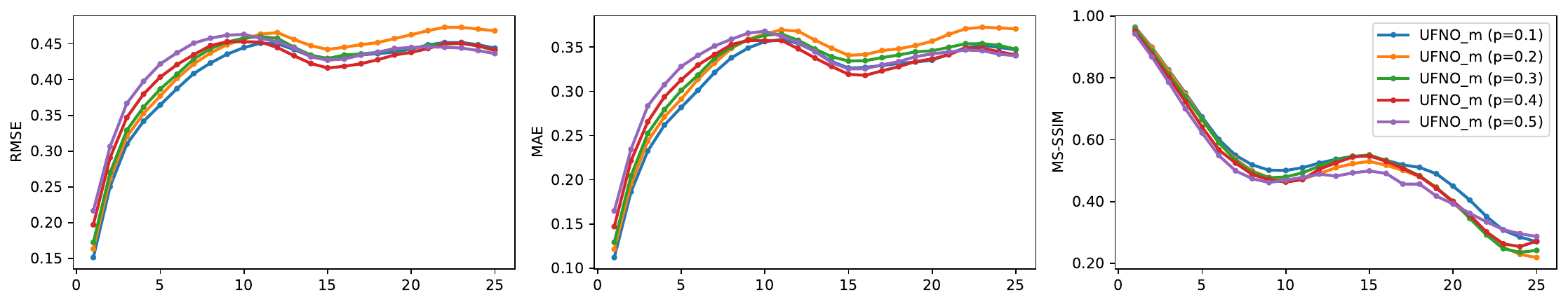}
        \caption{UFNO}
    \end{subfigure}
    \hfill 
    \begin{subfigure}{\textwidth}
        \centering
        \includegraphics[width=\textwidth]{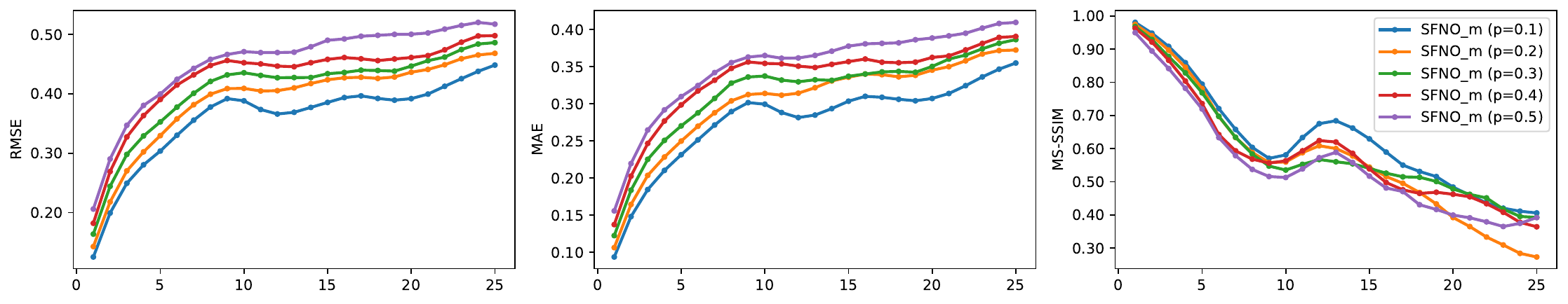}
        \caption{SFNO}
    \end{subfigure}
    
    \caption{Ablating the best strategies to yield the optimal ensembles for MC-dropout in Kolmogorov Flow, by varying dropout probability $p$ across baseline models. RMSE (\textit{left}), MAE (\textit{center}), MS-SSIM (\textit{right}).}
    \label{si-fig:kolmogorov_ablation_dropout}
\end{figure}

\noindent \textbf{Kolmogorov Flow}. Figure \ref{si-fig:kolmogorov_ablation_dropout} demonstrates that a dropout rate $p=0.1$ appears to yield the best ensembles across baseline models: TFNO, UFNO, SFNO. 

\begin{figure}[h]
    \centering
    \begin{subfigure}{\textwidth}
        \centering
        \includegraphics[width=\textwidth]{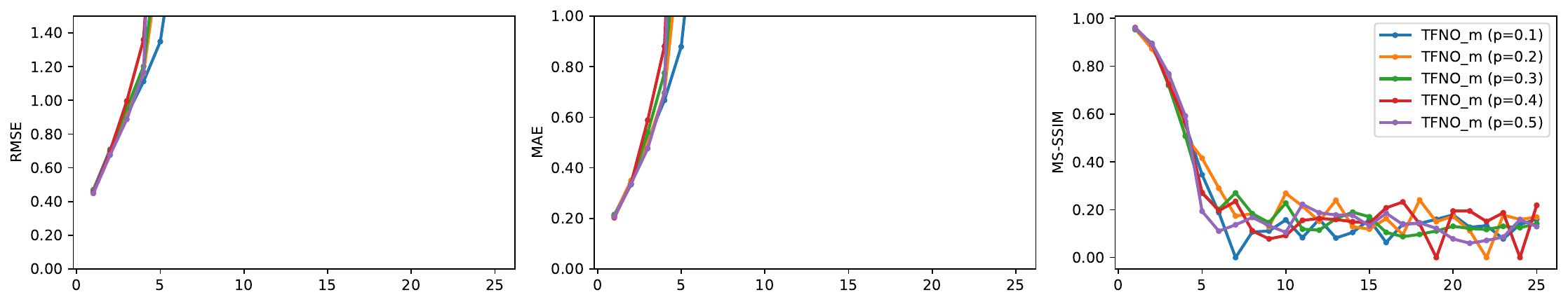}
        \caption{TFNO}
    \end{subfigure}
    \hfill
    \begin{subfigure}{\textwidth}
        \centering
        \includegraphics[width=\textwidth]{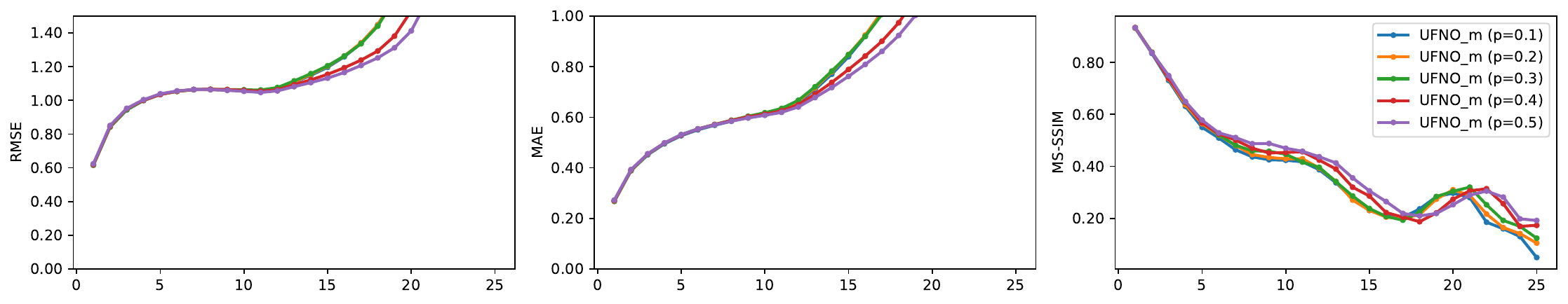}
        \caption{UFNO}
    \end{subfigure}
    \hfill 
    \begin{subfigure}{\textwidth}
        \centering
        \includegraphics[width=\textwidth]{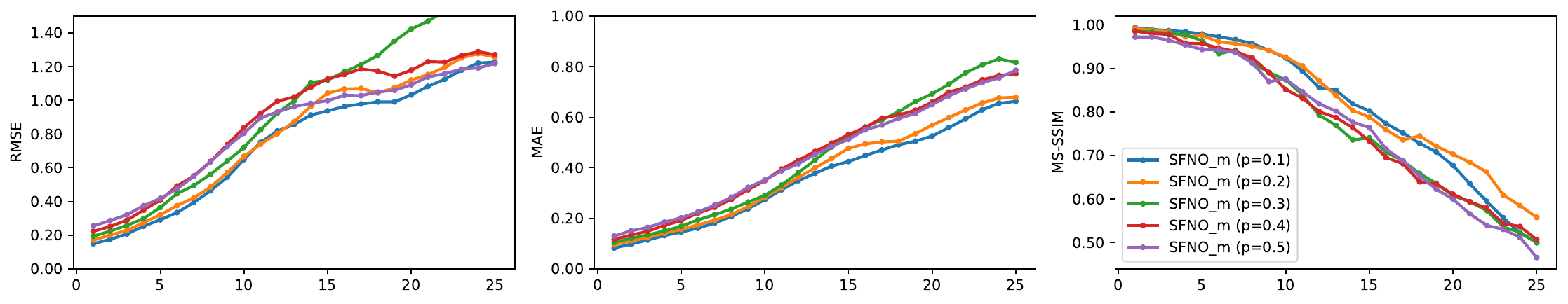}
        \caption{SFNO}
    \end{subfigure}
    
    \caption{Ablating the best strategies to yield the optimal ensembles for MC-dropout in Shallow Water Equation, by varying dropout probability $p$ across baseline models. RMSE (\textit{left}), MAE (\textit{center}), MS-SSIM (\textit{right}).}
    \label{si-fig:swe_ablation_dropout}
\end{figure}

\noindent \textbf{Shallow Water Equation}. Figure \ref{si-fig:swe_ablation_dropout} demonstrates that most baselines, especially TFNO and UFNO are extremely unstable over long-range forecasts. SFNO is the most stable one, and a dropout rate $p=0.1$ appears to yield the best ensembles. 

\clearpage

\subsection{Ensembling: IC perturbation}
We introduce Gaussian noise to initial condition (IC) following $\varepsilon \sim \mathcal{N}(0, f\mathbf{I})$, where $f \in [0,1]$ in order to construct an ensemble forecasts. \\

\begin{figure}[h]
    \centering
    \begin{subfigure}{\textwidth}
        \centering
        \includegraphics[width=\textwidth]{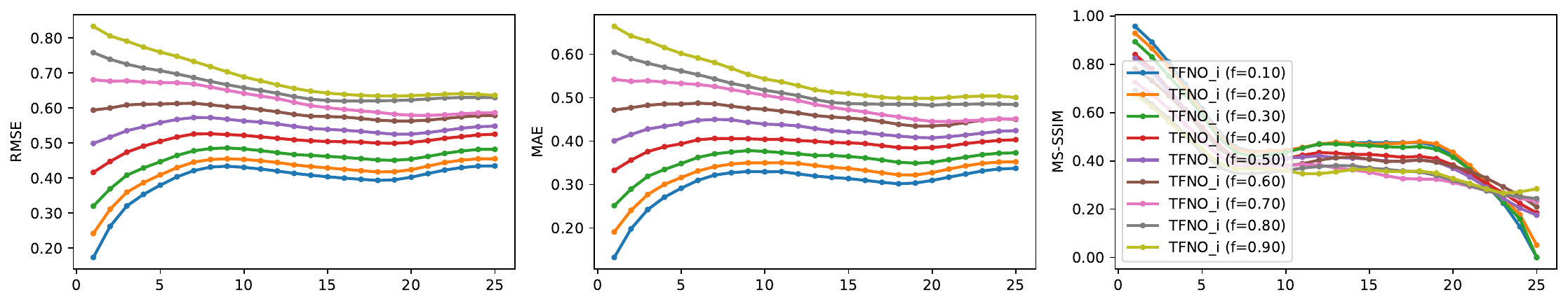}
        \caption{TFNO}
    \end{subfigure}
    \hfill
    \begin{subfigure}{\textwidth}
        \centering
        \includegraphics[width=\textwidth]{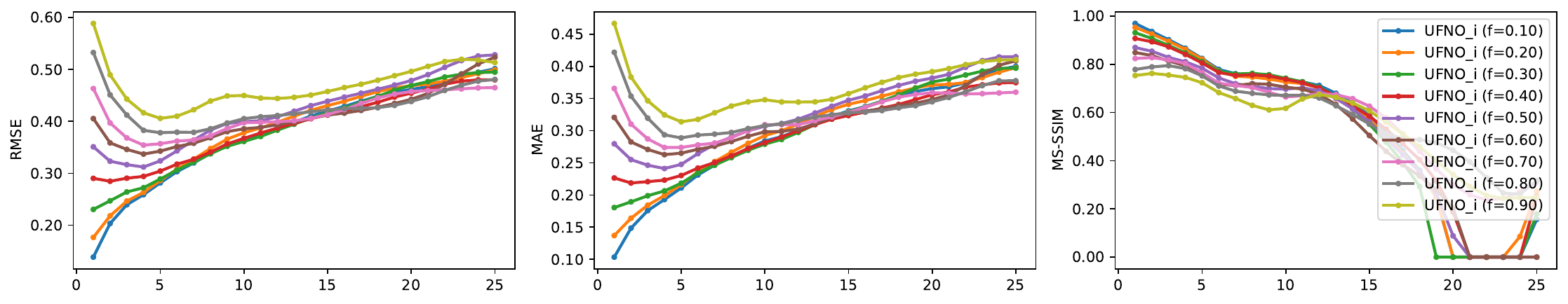}
        \caption{UFNO}
    \end{subfigure}
    \hfill 
    \begin{subfigure}{\textwidth}
        \centering
        \includegraphics[width=\textwidth]{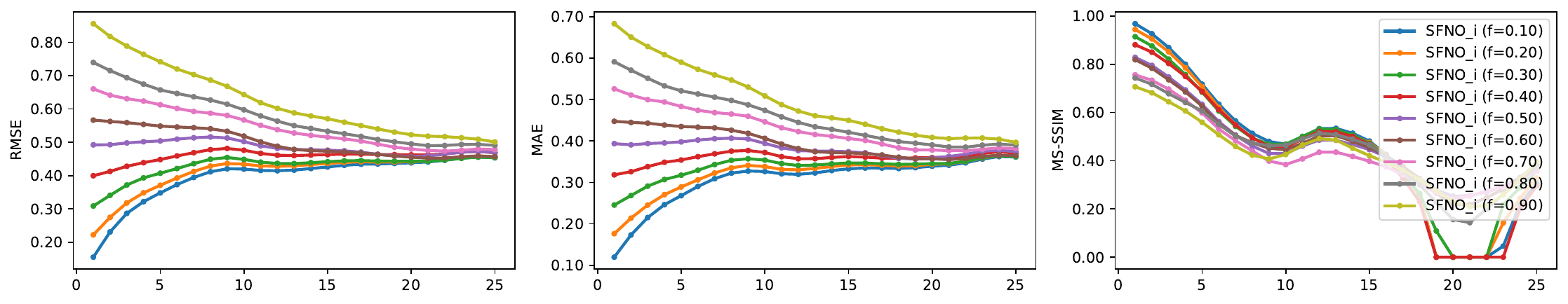}
        \caption{SFNO}
    \end{subfigure}
    
    \caption{Ablating the best strategies to yield the optimal ensembles for IC perturbation in Kolmogorov Flow, by varying Gaussian noise factor $f$ across baseline models. RMSE (\textit{left}), MAE (\textit{center}), MS-SSIM (\textit{right}).}
    \label{si-fig:kolmogorov_ablation_perturb}
\end{figure}

\noindent \textbf{Kolmogorov Flow}. Figure \ref{si-fig:kolmogorov_ablation_perturb} demonstrates that most baselines, especially TFNO and UFNO are extremely unstable over long-range forecasts. SFNO is the most stable one, and a dropout rate $f=0.1$ appears to yield the best ensembles. 

\begin{figure}[h]
    \centering
    \begin{subfigure}{\textwidth}
        \centering
        \includegraphics[width=\textwidth]{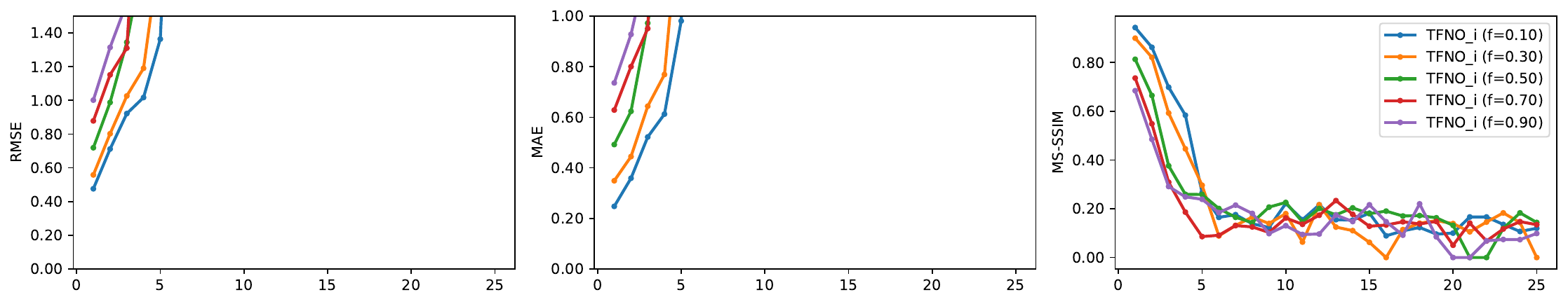}
        \caption{TFNO}
    \end{subfigure}
    \hfill
    \begin{subfigure}{\textwidth}
        \centering
        \includegraphics[width=\textwidth]{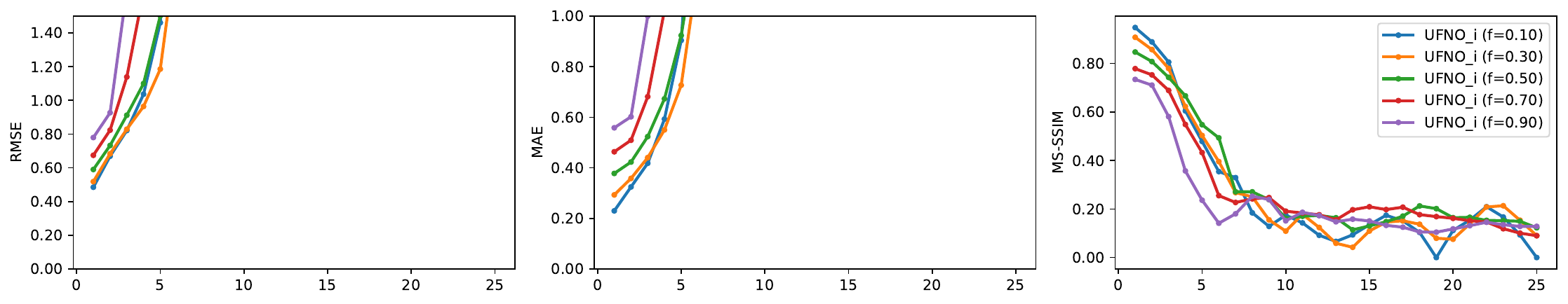}
        \caption{UFNO}
    \end{subfigure}
    \hfill 
    \begin{subfigure}{\textwidth}
        \centering
        \includegraphics[width=\textwidth]{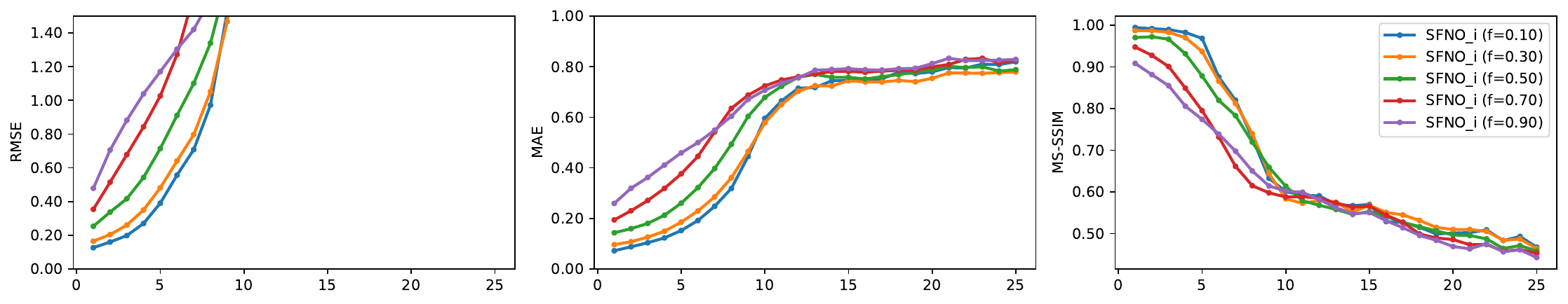}
        \caption{SFNO}
    \end{subfigure}
    
    \caption{Ablating the best strategies to yield the optimal ensembles for IC perturbation in Shallow Water Equation, by varying Gaussian noise factor $f$ across baseline models. RMSE (\textit{left}), MAE (\textit{center}), MS-SSIM (\textit{right}).}
    \label{si-fig:swe_ablation_perturb}
\end{figure}

\noindent \textbf{Shallow Water Equation}. Figure \ref{si-fig:swe_ablation_perturb} demonstrates that a perturbation factor $f=0.1$ appears to yield the best ensembles across baseline models: TFNO, UFNO, SFNO. 

\clearpage

\subsection{Physical consistency checks}
\begin{figure}[h]
    \begin{subfigure}{0.45\textwidth}
        \centering
        \includegraphics[width=\textwidth]{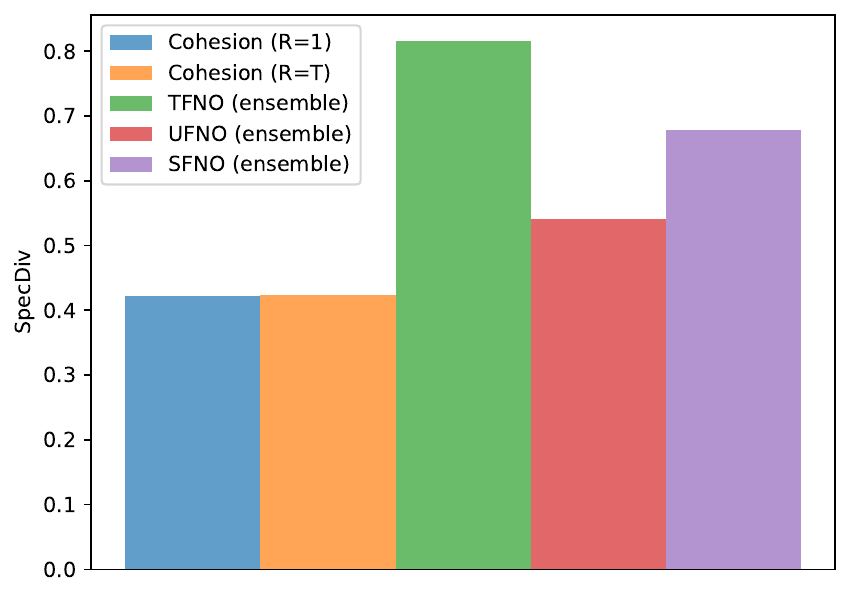}
        \caption{Kolmogorov Flow}
    \end{subfigure}
    \hfill
    \begin{subfigure}{0.45\textwidth}
        \centering
        \includegraphics[width=\textwidth]{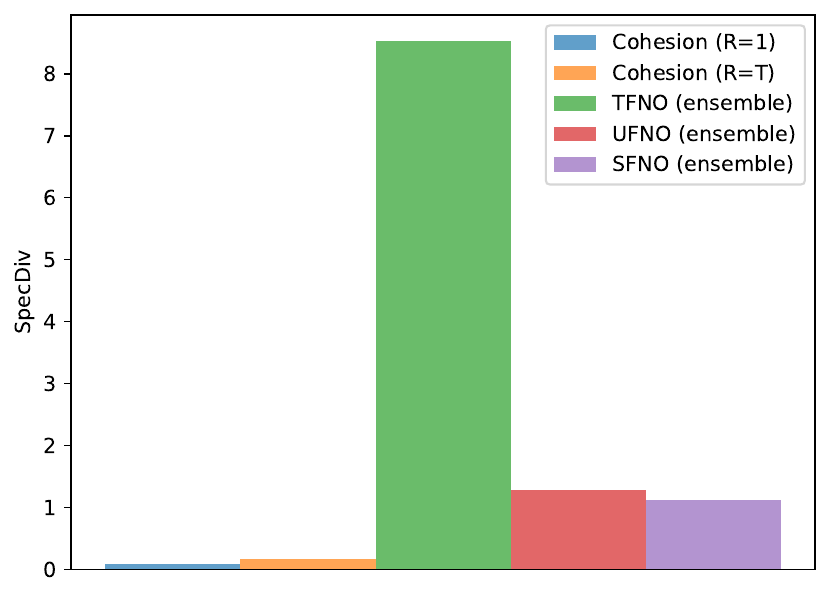}
        \caption{Shallow Water Equation}
    \end{subfigure}

    \caption{Cohesion has the lowest spectral divergence ($\downarrow$) across probabilistic baselines, indicating its ability to capture multiscale physical structures.}
    \label{si-fig:specdiv}
\end{figure}

We provide more results for physical consistency experiments introduced in the main text. Figure \ref{si-fig:specdiv} highlights the ability of Cohesion to capture multiscale physics by having the lowest spectral divergence across probabilistic baselines.

\begin{figure}[h]
    \centering
    \begin{subfigure}{\textwidth}
        \centering
        \includegraphics[width=\textwidth]{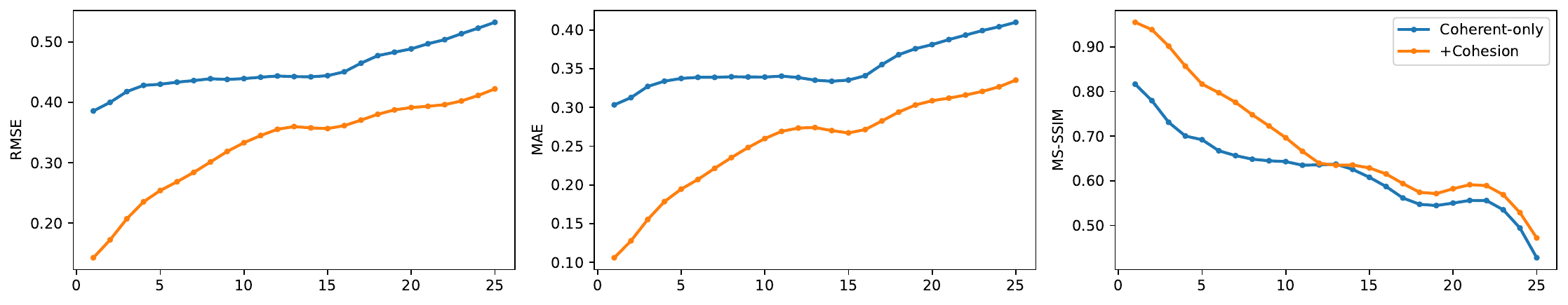}
        \caption{Kolmogorov Flow}
    \end{subfigure}
    \hfill
    \begin{subfigure}{\textwidth}
        \centering
        \includegraphics[width=\textwidth]{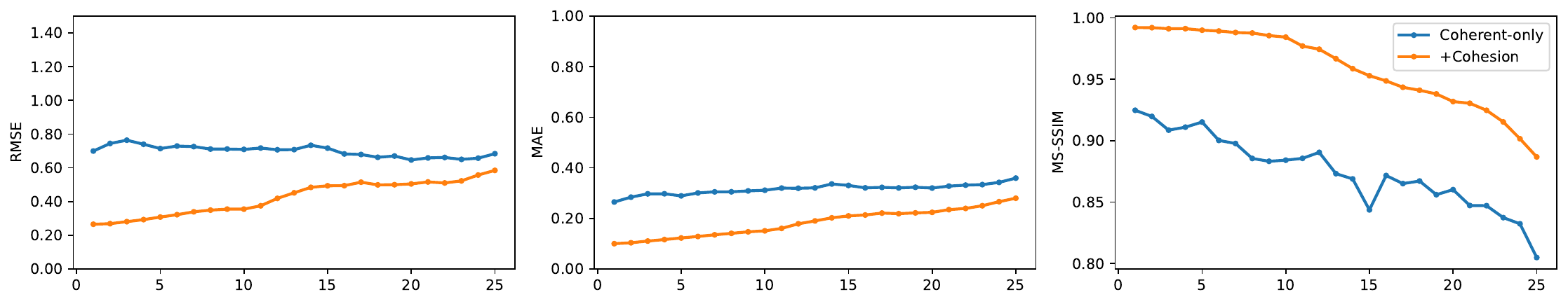}
        \caption{Shallow Water Equation}
    \end{subfigure}
    \hfill 
    
    \caption{Cohesion improves RMSE ($\downarrow$), MAE ($\downarrow$), and MS-SSIM ($\uparrow$) scores over its coherent-only prior forecasts generated sequentially with ROMs.}
    \label{si-fig:ablation_refiner}
\end{figure}

Figure \ref{si-fig:ablation_refiner} demonstrates that Cohesion not only improves spectral metrics, but in so doing also boost the more traditional non-spectral scores.

\subsection{Scalability of coherent estimator}
\begin{figure}[h]
    \centering
    \begin{subfigure}{\textwidth}
        \centering
        \includegraphics[width=\textwidth]{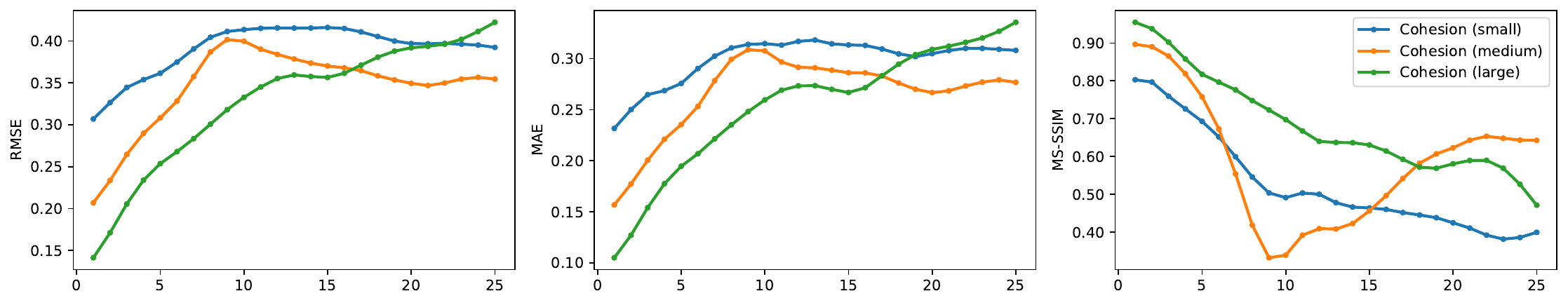}
        \caption{Kolmogorov Flow}
    \end{subfigure}
    \hfill
    \begin{subfigure}{\textwidth}
        \centering
        \includegraphics[width=\textwidth]{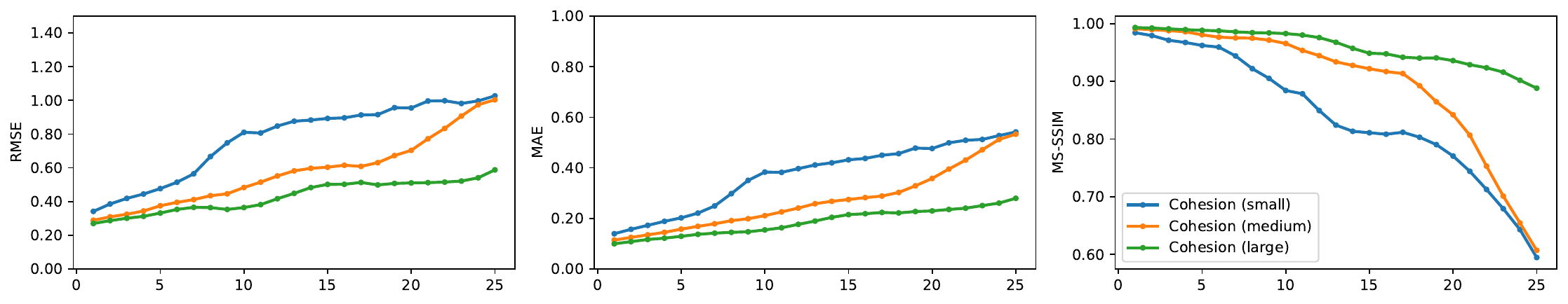}
        \caption{Shallow Water Equation}
    \end{subfigure}
    
    \caption{Ablating the scaling properties of coherent estimators. RMSE (\textit{left}), MAE (\textit{center}), MS-SSIM (\textit{right}).}
    \label{si-fig:ablation_scaling}
\end{figure}

One interesting experiment is to check the scaling of our coherent estimators. In the trajectory planning setting (R=T), we find that increasing the number of parameters in the coherent estimators help to improve the quantitative results, but plateaus given our best setting ("large"), as generally illustrated in Figure \ref{si-fig:ablation_scaling}.

\subsection{Choice of coherent estimators}
\begin{figure}[h]
    \centering
    \begin{subfigure}{\textwidth}
        \centering
        \includegraphics[width=\textwidth]{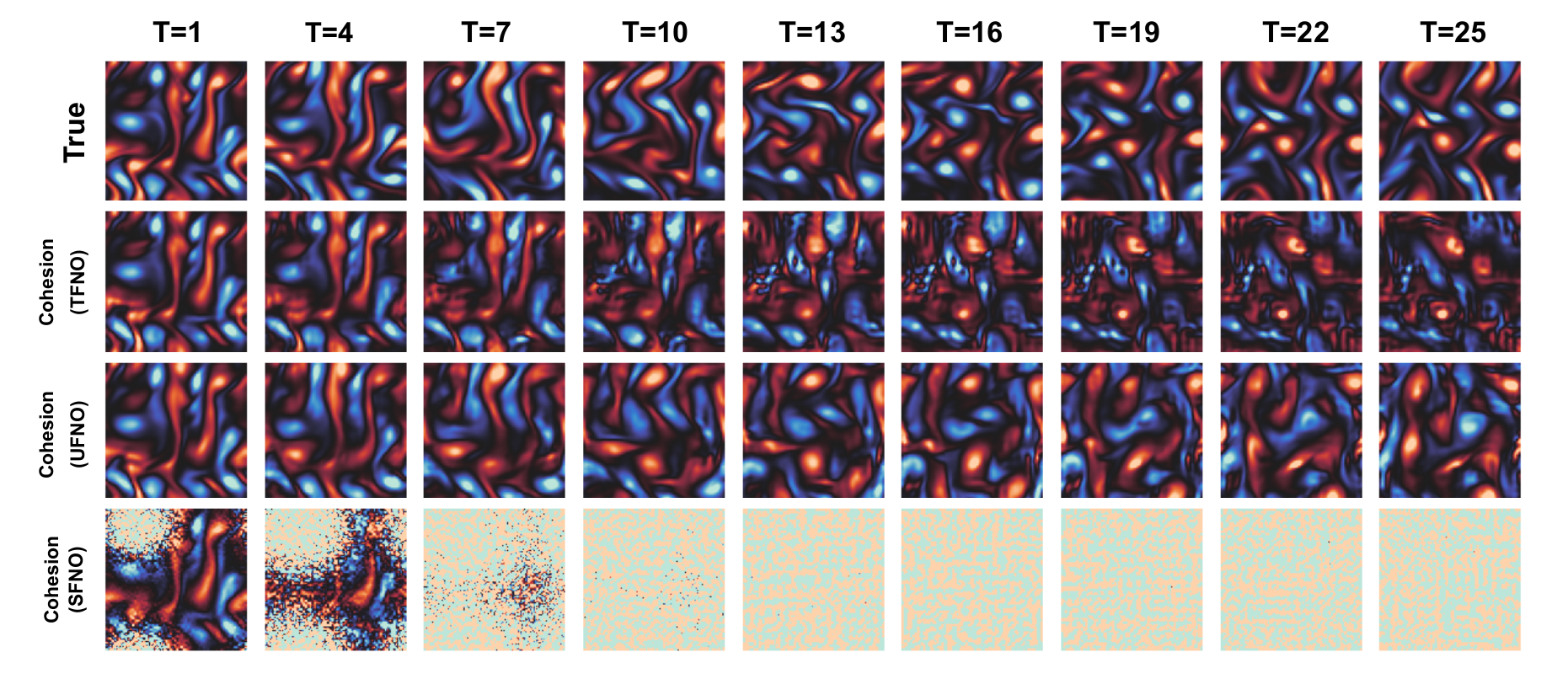}
        \caption{Kolmogorov Flow}
    \end{subfigure}
    \hfill
    \begin{subfigure}{\textwidth}
        \centering
        \includegraphics[width=\textwidth]{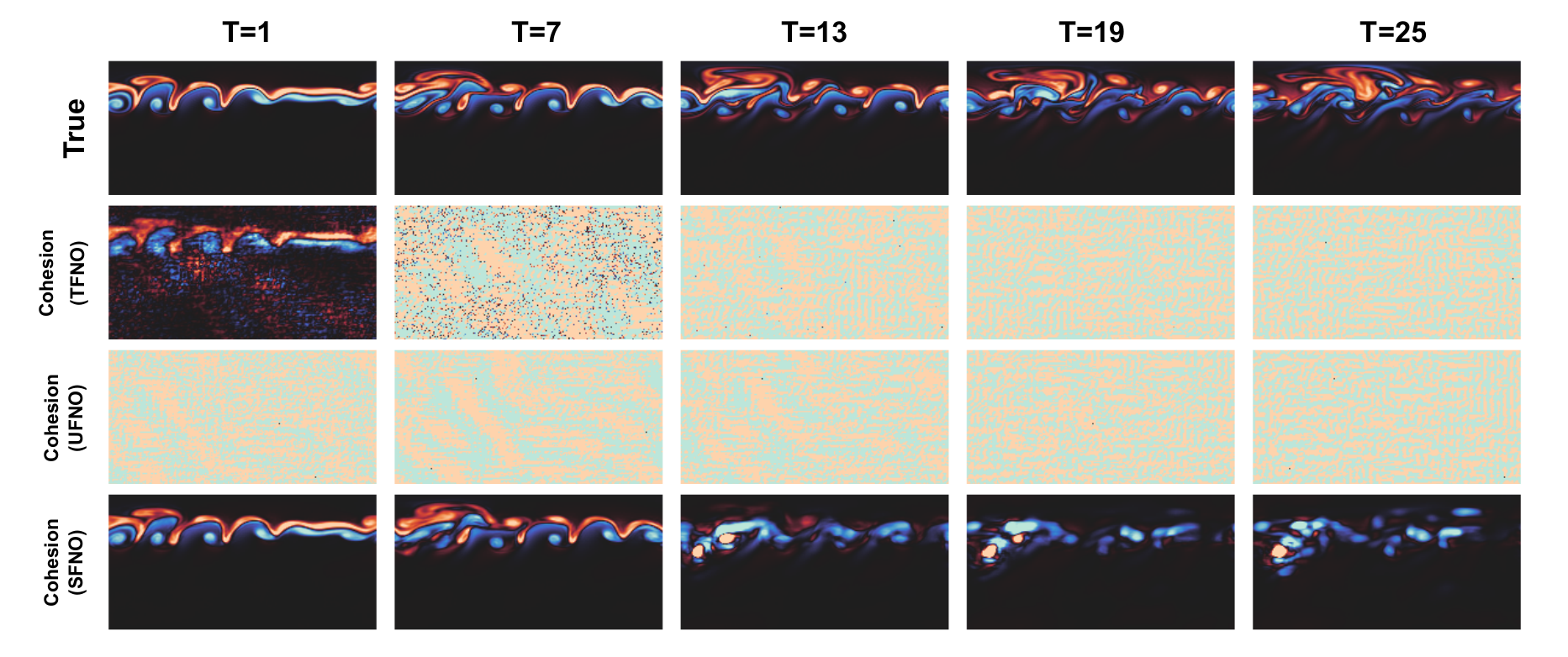}
        \caption{Shallow Water Equation}
    \end{subfigure}
    
    \caption{Qualitative single-member realization where Cohesion is guided by high-dimensional coherent estimators, including TFNO, UFNO, and SFNO.}
    \label{si-fig:ablation_prior}
\end{figure}

We further investigate the effect of using different coherent prior estimators within Cohesion. As shown in Figure~\ref{si-fig:ablation_prior}, replacing ROMs with high-dimensional priors—such as TFNO, UFNO, and SFNO—generally degrades generative performance. This observation reinforces our initial hypothesis that diffusion-based models are highly sensitive to the quality of the conditioning prior. 

\clearpage

\section{Additional Results}
\begin{figure}[h]
    \centering
    \includegraphics[width=\textwidth]{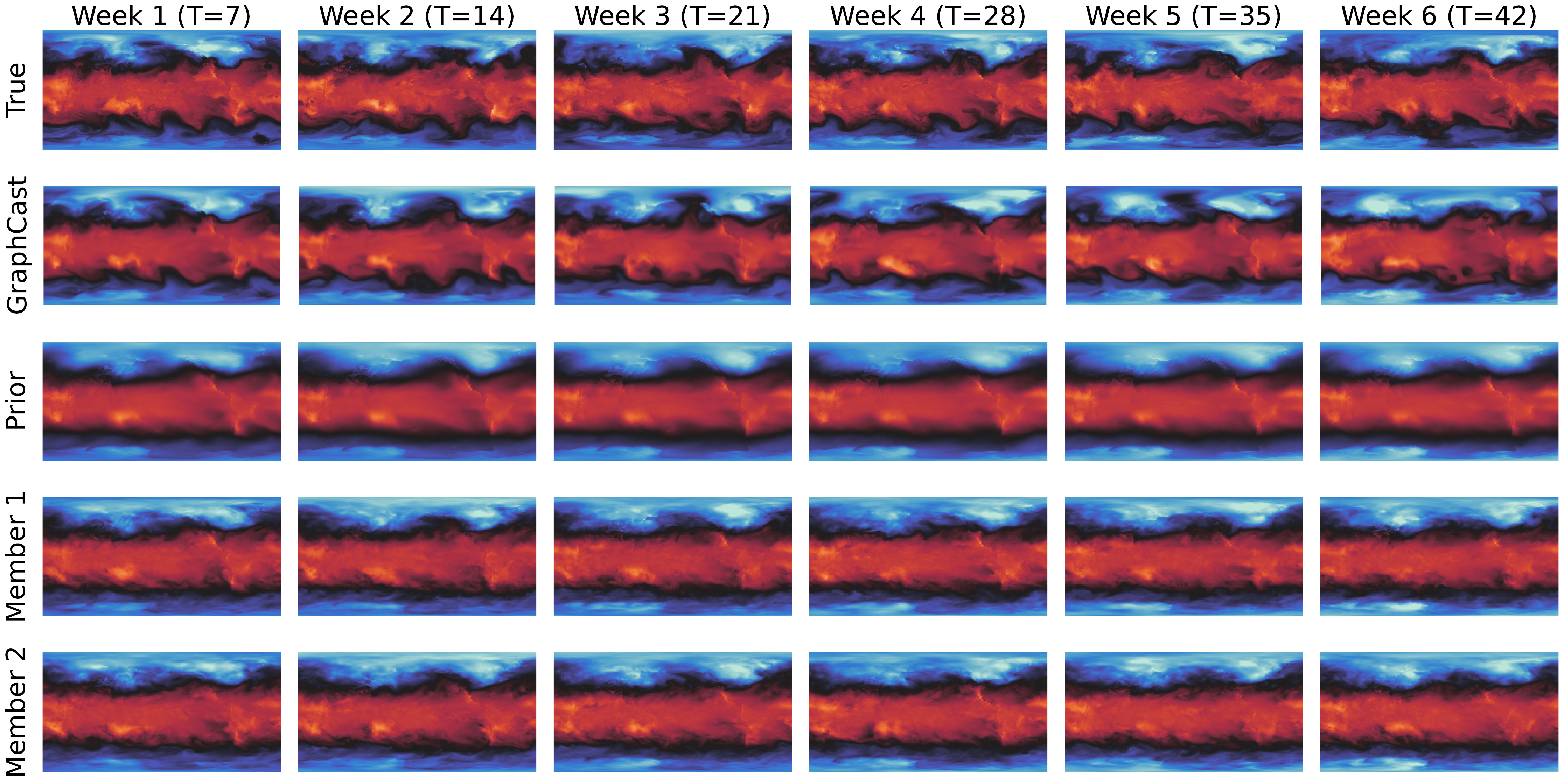}
    \caption{Qualitative realization of S2S dynamics ($t850$), where Cohesion improves the fidelity of forecasts over smoothed-out priors. Initialization date: $2022$-$01$-$01$; results are shown for the simulation snapshot at each of the 6 weeks. We additionally include GraphCast as representative of existing DLWP to highlight the smoothing effect.}
    \label{si-fig:s2s_t850}
\end{figure}

\begin{figure}[h]
    \centering
    \includegraphics[width=\textwidth]{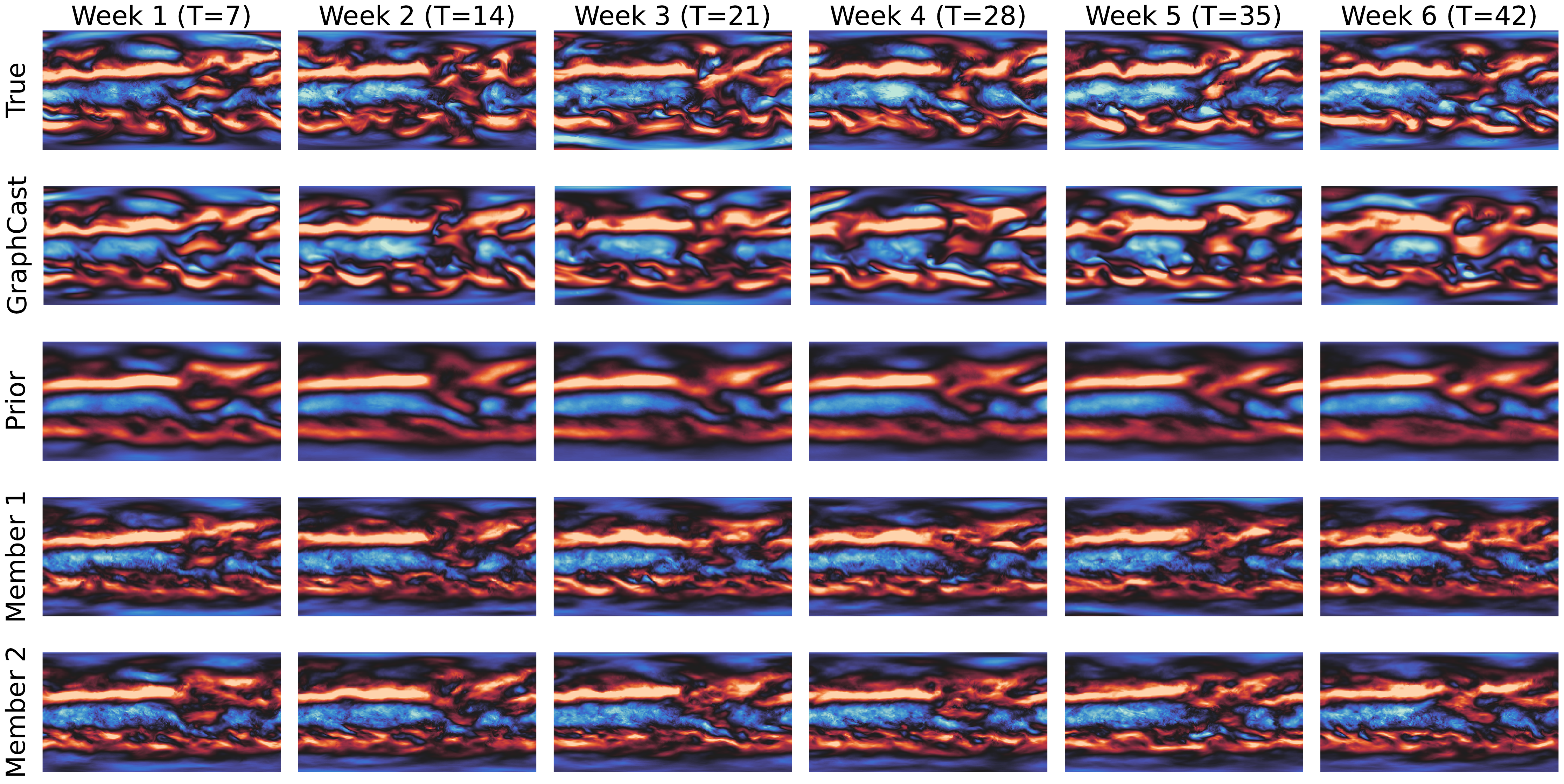}
    \caption{Qualitative realization of S2S dynamics ($u200$), where Cohesion improves the fidelity of forecasts over smoothed-out priors. Initialization date: $2022$-$01$-$01$; results are shown for the simulation snapshot at each of the 6 weeks. We additionally include GraphCast as representative of existing DLWP to highlight the smoothing effect.}
    \label{si-fig:s2s_u200}
\end{figure}

\begin{figure}[h]
    \centering
    \includegraphics[width=\textwidth]{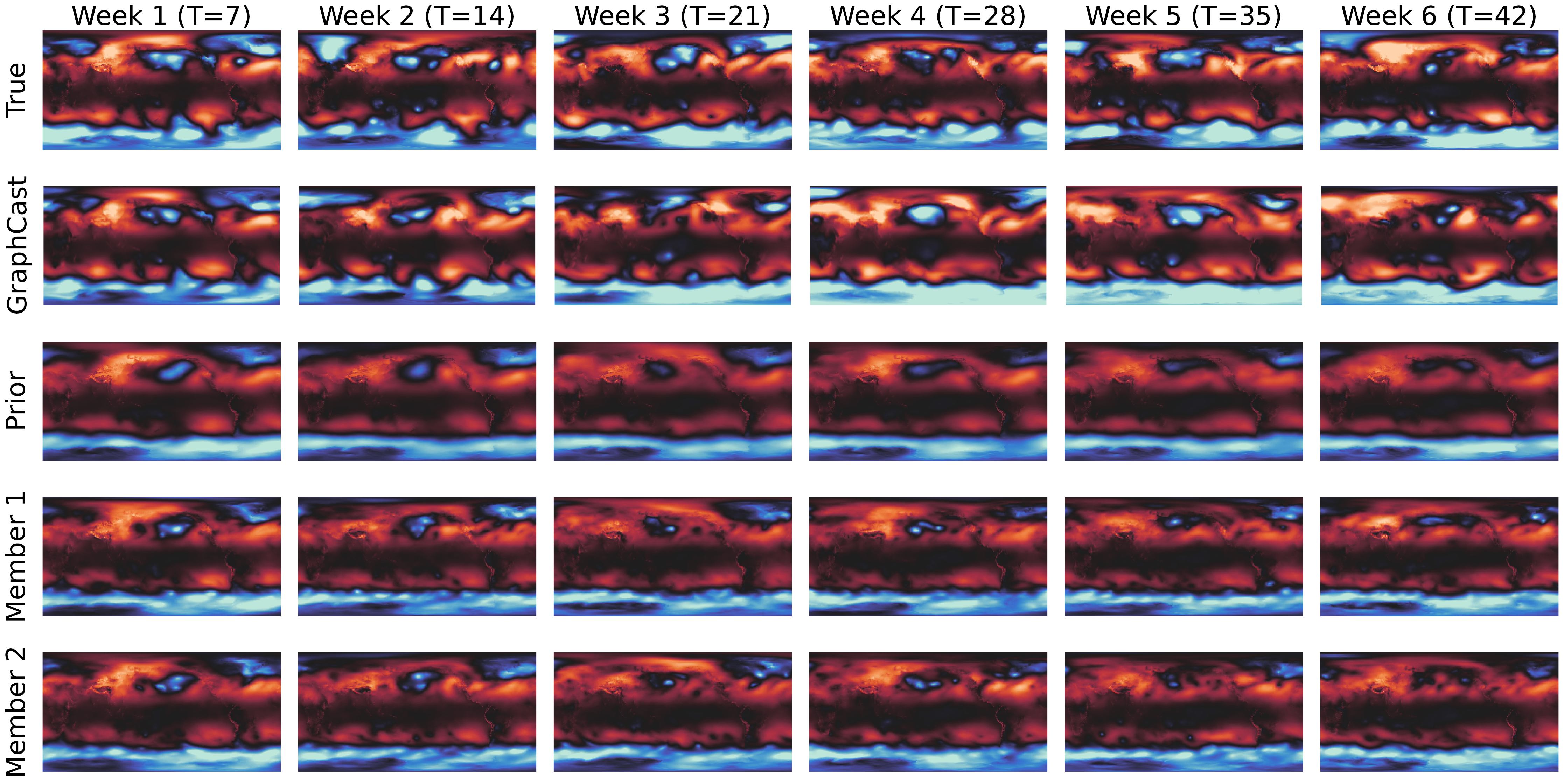}
    \caption{Qualitative realization of S2S dynamics ($z1000$), where Cohesion improves the fidelity of forecasts over smoothed-out priors. Initialization date: $2022$-$01$-$01$; results are shown for the simulation snapshot at each of the 6 weeks. We additionally include GraphCast as representative of existing DLWP to highlight the smoothing effect.}
    \label{si-fig:s2s_z1000}
\end{figure}

\begin{figure}[h!]
    \centering
    \begin{subfigure}{\textwidth}
        \centering
        \includegraphics[width=\textwidth]{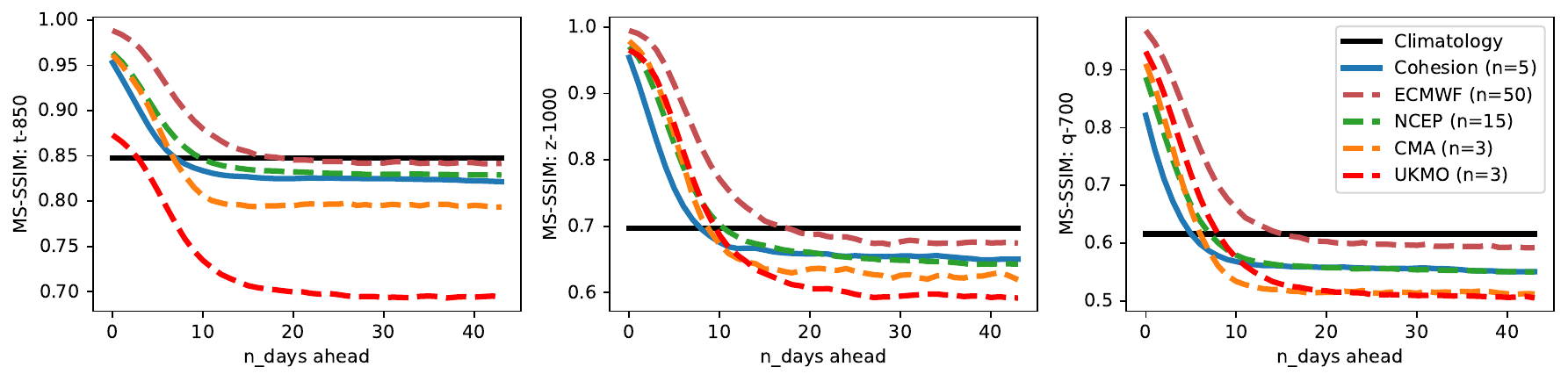}
        \caption{MS-SSIM ($\uparrow$)}
    \end{subfigure}
    \hfill 
    \begin{subfigure}{\textwidth}
        \centering
        \includegraphics[width=\textwidth]{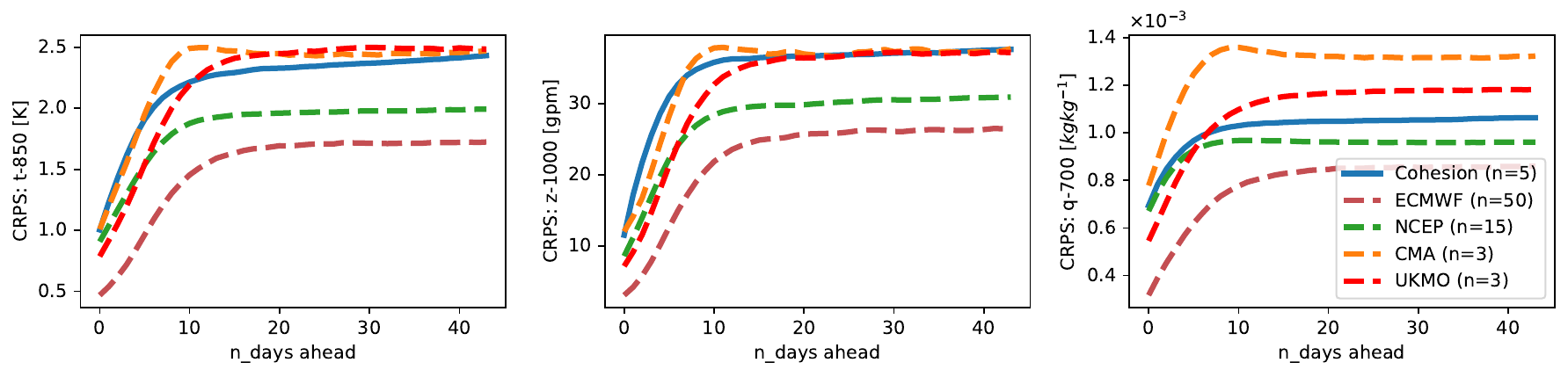}
        \caption{CRPS ($\downarrow$)}
    \end{subfigure}
    
    \caption{Cohesion is competitive with the ensemble forecasts from ECMWF and NCEP, while often outperforming CMA and UKMO.}
    \label{si-fig:s2s_rollout_metrics}
\end{figure}

\begin{figure}[h]
    \centering 
    \begin{subfigure}{\textwidth}
        \centering
        \includegraphics[width=0.8\textwidth]{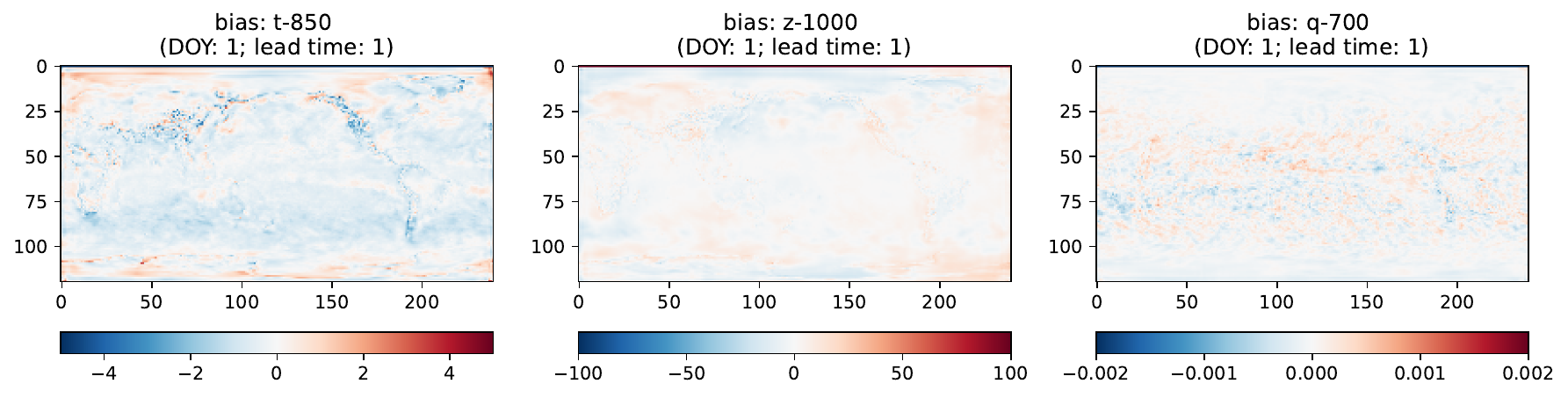}
        \caption{DOY = 2}
    \end{subfigure}
    \hfill
    \begin{subfigure}{\textwidth}
        \centering
        \includegraphics[width=0.8\textwidth]{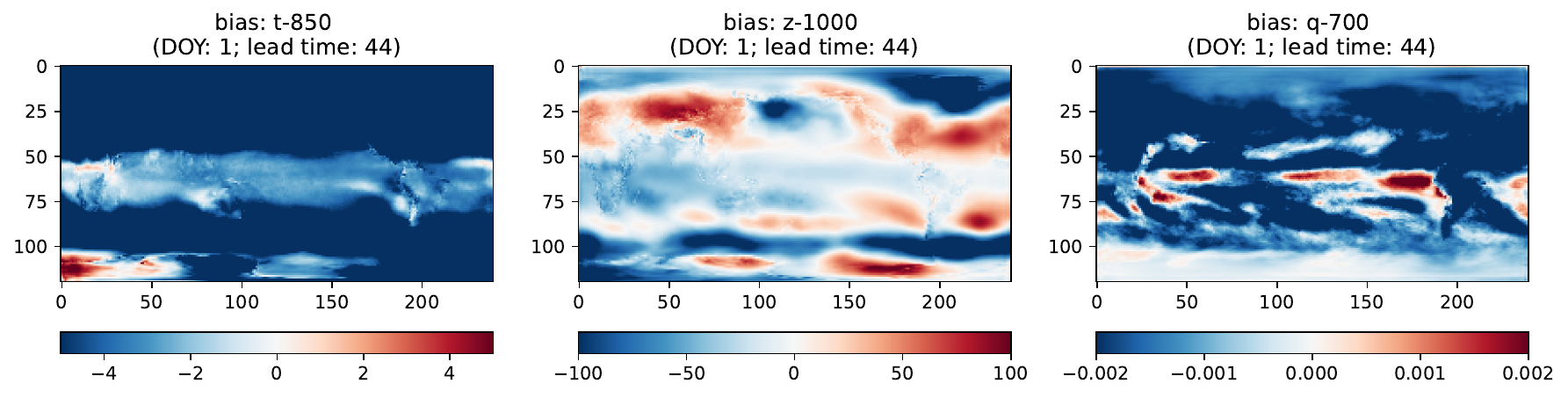}
        \caption{DOY = 45}
    \end{subfigure}
    \caption{Mean spatially varying bias of our coherent estimators across different day-of-year (DOY), used to debias the priors for S2S emulation at inference time. As expected, bias drift increases with lead time.}
    \label{si-fig:s2s_bias_correct}
\end{figure}

We provide additional results, including Figure \ref{si-fig:s2s_t850} - \ref{si-fig:s2s_z1000} for more qualitative evaluations on S2S. Figure~\ref{si-fig:s2s_rollout_metrics} for S2S evaluation metrics, and Figure~\ref{si-fig:s2s_bias_correct} for qualitative visualization of bias accumulation in the coherent estimators over long-range forecasts.

\end{appendices}

\end{document}